\documentclass{article}
\usepackage{arxiv, quickmath}

\usepackage[utf8]{inputenc} 
\usepackage[T1]{fontenc}    
\usepackage{hyperref}       
\usepackage{url}            
\usepackage{booktabs}       
\usepackage{amsfonts}       
\usepackage{nicefrac}       
\usepackage{microtype}      
\usepackage{xcolor}         
\usepackage{graphicx}
\usepackage{subfigure}
\usepackage{enumitem}
\usepackage{adjustbox}
\usepackage{wrapfig}
\usepackage{tabularx}
\usepackage[normalem]{ulem}
\usepackage{multirow}
\usepackage{svg}
\usepackage{algorithm}
\usepackage{algorithmic}

\usetikzlibrary{arrows.meta, positioning}

\usepackage{rotating}
\usepackage{multirow}
\usepackage{makecell}
\usepackage{tabularray}

\graphicspath{{./figures/}}

\usepackage{array}
\newcolumntype{L}[1]{>{\raggedright\let\newline\\\arraybackslash\hspace{0pt}}m{#1}}
\newcolumntype{C}[1]{>{\centering\let\newline\\\arraybackslash\hspace{0pt}}m{#1}}
\newcolumntype{R}[1]{>{\raggedleft\let\newline\\\arraybackslash\hspace{0pt}}m{#1}}

\usepackage{environ}
\NewEnviron{casestudy}[1]{
    \setlength{\leftskip}{0.5cm}
    
    \emph{Case Study: #1.} \BODY
    
    \setlength{\leftskip}{0cm}
}

\title{Understanding the Theoretical Foundations of Deep Neural Networks through Differential Equations}
\author{
    Hongjue Zhao \\
    University of Illinois Urbana Champaign \\ 
\And
    Yizhuo Chen \\
    University of Illinois Urbana Champaign \\
\And
    Yuchen Wang \\
    William \& Mary \\
\And
    Hairong Qi \\
    University of Tennessee, Knoxville \\
\And 
    Lui Sha \\ 
    University of Illinois Urbana Champaign \\
\And
    Tarek Abdelzaher \thanks{Corresponding authors.}\\
    University of Illinois Urbana Champaign \\
\And
    Huajie Shao \footnotemark[1]\\
    William \& Mary
}
\date{}

\begin{document}
\maketitle

\renewcommand{\arraystretch}{1.5}
\begin{abstract}
    Deep neural networks (DNNs) have achieved remarkable empirical success, yet the absence of a principled theoretical foundation continues to hinder their systematic development. In this survey, we present differential equations as a theoretical foundation for understanding, analyzing, and improving DNNs. We organize the discussion around three guiding questions: \textit{i)} how differential equations offer a principled understanding of DNN architectures, \textit{ii)} how tools from differential equations can be used to improve DNN performance in a principled way, and \textit{iii)} what real-world applications benefit from grounding DNNs in differential equations. We adopt a two-fold perspective spanning the \textit{model level}, which interprets the whole DNN as a differential equation, and the \textit{layer level}, which models individual DNN components as differential equations. From these two perspectives, we review how this framework connects model design, theoretical analysis, and performance improvement. We further discuss real-world applications, as well as key challenges and opportunities for future research.
\end{abstract}


\section{Introduction}\label{sec:intro}

Deep Neural Networks (DNNs), as powerful universal function approximators, have driven transformative breakthroughs across a wide range of scientific and technological domains over the past decade. Their impact spans computer vision~\cite{krizhevsky2012imagenet, dosovitskiy2020image}, natural language processing~\cite{vaswaniAttentionAllYou2017, touvron2023llama}, reinforcement learning~\cite{arulkumaran2017deep, jamshidi2025application}, generative modeling~\cite{Song:2020hus, LipmanFlowMatchingGuide2024}, and scientific discovery~\cite{jumper2021highly}, among others. Collectively, these advances have firmly established DNNs as foundational tools in modern artificial intelligence.

Despite their remarkable empirical success, the absence of a principled theoretical foundation continues to hinder the systematic development of DNNs, leading to two major consequences. On the one hand, the design of novel architectures remains largely heuristic. For example, some of the most widely adopted architectures, including convolutional neural networks (CNNs)~\cite{krizhevsky2012imagenet, he2016deep}, recurrent neural networks (RNNs)~\cite{hochreiter1997long, cho2014learning}, and Transformers~\cite{vaswaniAttentionAllYou2017}, were originally proposed based on distinct intuitions and design heuristics, rather than grounded in a unified theoretical principle. On the other hand, improvements to existing architectures are often motivated by empirical observations rather than rigorous analysis. Notable examples include skip connections in residual networks~\cite{he2016deep} and gating mechanisms in RNNs~\cite{hochreiter1997long, cho2014learning}, both of which were introduced primarily through empirical insights rather than derived from first principles.

In response to these limitations, an emerging body of research has explored reinterpretations of DNNs through the lens of classical mathematical theory. Among the various mathematical frameworks considered, \emph{differential equations} have emerged as a particularly powerful and unifying perspective --- one that views DNNs, in many cases, as \textit{discretized differential equations}~\cite{EProposalMachineLearning2017, ChenNeuralOrdinaryDifferential2018a, kidgerNeuralDifferentialEquations2022, GuMambaLinearTimeSequence2023}. One of the earliest and most influential insights in this direction showed that residual networks (ResNets)~\cite{he2016deep} can be interpreted as \emph{forward Euler discretizations} of ordinary differential equations (ODEs)~\cite{EProposalMachineLearning2017, haberStableArchitecturesDeep2018}. Building on this connection, researchers proposed a variety of architectures grounded in alternative numerical schemes, aiming to improve expressiveness or stability~\cite{ZhangPolyNetPursuitStructural2017, LarssonFractalNetUltraDeepNeural2017, GomezReversibleResidualNetwork2017, luFiniteLayerNeural2018a}. These developments culminated in the introduction of \emph{Neural ODEs}~\cite{ChenNeuralOrdinaryDifferential2018a}, in which the forward pass of a DNN is defined by solving an ODE with a numerical solver, thereby enabling continuous-depth models and adaptive computation. Subsequent research extended the idea beyond standard ODEs to include controlled differential equations (CDEs)~\cite{KidgerNeuralControlledDifferential2020, MorrillNeuralRoughDifferential2021} and stochastic differential equations (SDEs)~\cite{tzen2019neural, jia2019neural, KidgerNeuralSDEsInfiniteDimensional2021}, giving rise to the broader class of \emph{Neural Differential Equations} (NDEs)~\cite{kidgerNeuralDifferentialEquations2022}. Based on this class, flow-based generative models, such as flow models~\cite{ChenNeuralOrdinaryDifferential2018a, LipmanFlowMatchingGuide2024} and diffusion models~\cite{ho2020denoising}, can be interpreted as instances of either deterministic or stochastic differential equations~\cite{Song:2020hus, LipmanFlowMatchingGuide2024, albergo2023stochastic}, where sampling corresponds to numerically simulating the underlying dynamical system.

In addition to viewing the entire architecture as a differential equation, a complementary line of work focuses on constructing individual \emph{layers} within DNNs as differential equations, leading to the development of \emph{Deep State Space Models} (Deep SSMs) \cite{GuCombiningRecurrentConvolutional2021a, GuEfficientlyModelingLong2022, GuMambaLinearTimeSequence2023, PatroMamba360SurveyState2024}. This line of work originates from the High-order Polynomial Projection Operators (HiPPO) framework~\cite{GuHiPPORecurrentMemory2020}, which introduces a memory mechanism for long-range sequence modeling based on online function approximation. The core idea is to project a continuous-time input signal onto a series of orthogonal polynomial basis and represent it via its time-varying coefficients. Crucially, HiPPO demonstrates that these coefficients can be updated online through a \textit{linear state space model}, which is a kind of special ODE. This forms the foundation for Deep SSMs. Recent studies have revealed explicit connections between Deep SSMs and other fundamental DNN components, such as convolution, recurrence, and attention mechanisms~\cite{GuCombiningRecurrentConvolutional2021a, DaoTransformersAreSSMs2024, sieber2024understanding}. As a result, Deep SSMs offer a unified modeling paradigm that integrates the strengths of these architectures, such as the efficiency of convolution, the memory capacity of recurrence, and the expressivity of attention mechanism.

Building on the architectural reinterpretation enabled by differential equations, researchers have also begun to exploit the accompanying mathematical tools to analyze and improve DNNs. Fundamental properties of differential equations (e.g., stability, controllability, and expressivity, etc.) have been extensively investigated in applied mathematics. When DNNs are reformulated as differential equation systems, these well-established concepts offer a principled lens through which to study their behavior. Within this framework, desirable properties from differential equations can be deliberately imposed to target specific aspects of DNN performance, thereby providing theoretically grounded design and analysis tools. For example, Lyapunov-based stability criteria and barrier certificates have been employed to enhance robustness and interpretability, opening promising directions for certifiable learning and safe deployment~\cite{RodriguezLyaNetLyapunovFramework2022, YangCertifiablyRobustNeural2023}.

Taken together, these research directions demonstrate that viewing DNNs through the lens of differential equations offers a principled framework for addressing several core challenges:

\begin{enumerate}
    \item[\textit{(i)}] \emph{Principled architecture design.} Many modern DNN components can be interpreted as specific discretizations or formulations of differential equations, encompassing widely used layers such as attention, convolutional, and recurrent modules, as well as full architectures including ResNets, NDEs and diffusion models. This correspondence enables the systematic development of novel and effective neural architectures by leveraging theoretical tools from areas such as control theory, dynamical systems, and numerical analysis.
    
    \item[\textit{(ii)}] \emph{Performance improvement.} Mathematical tools from differential equations provide a structured means to analyze key properties of DNNs. By enforcing these desirable properties within network design or training, one can target and enhance specific aspects of DNN performance in a theoretically grounded manner.
\end{enumerate}

\begin{figure}[tbp]
    \centering
    \includegraphics[width=\textwidth]{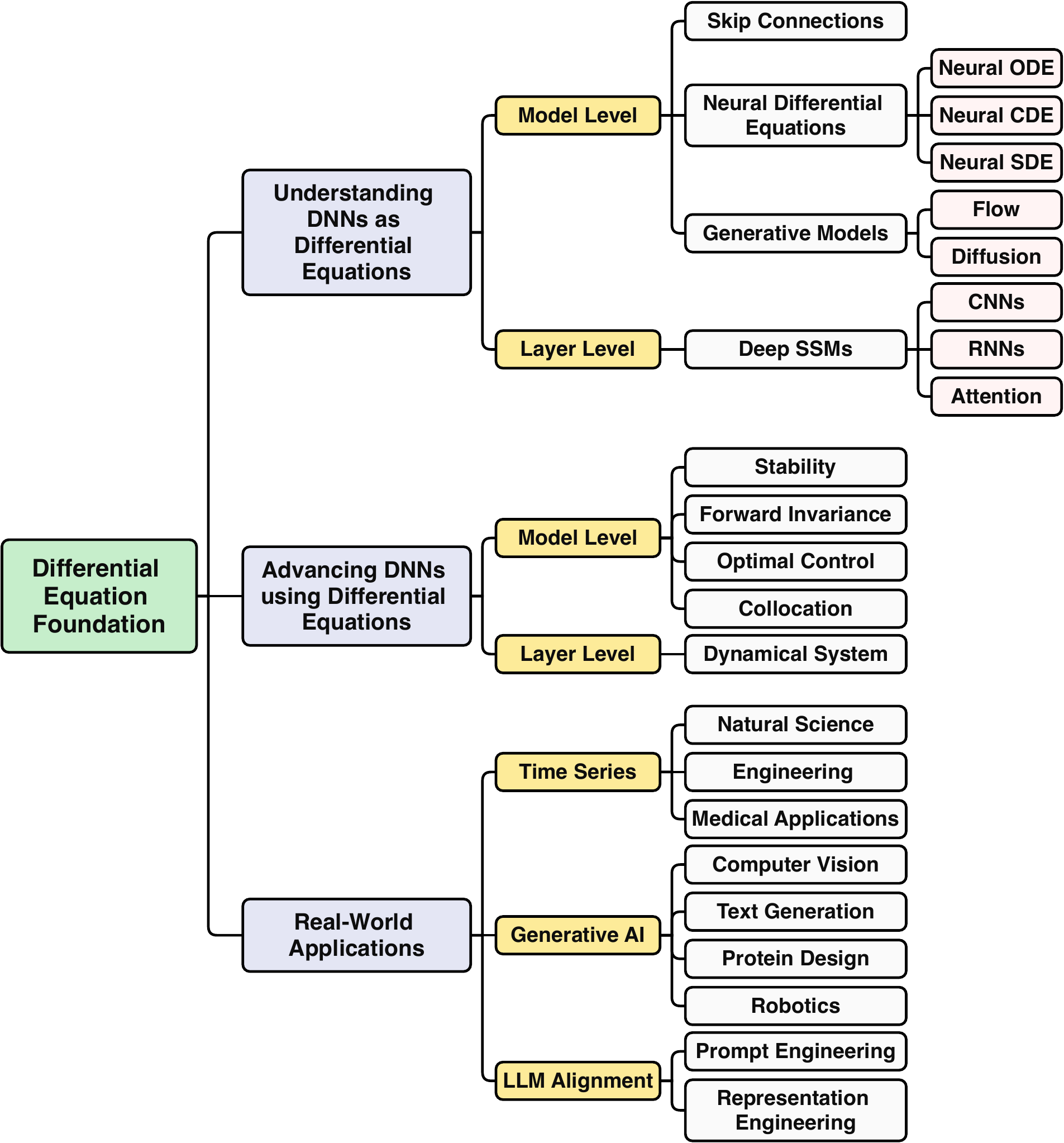}
    \caption{\textbf{Overview of the survey.} This survey is organized around three key questions concerning DNNs through the lens of differential equations: 
    \textit{(i)} How can differential equations offer a principled understanding of DNN architectures? 
    \textit{(ii)} How can tools from differential equations be used to improve DNN performance in a principled way? and 
    \textit{(iii)} What real-world applications benefit from grounding DNNs in differential equations?
    We focus on two levels of abstraction: the \emph{model level}, which interprets the entire DNN as a dynamical system, and the \emph{layer level}, which models individual layers of DNNs as differential equations.}
    \label{fig:overview}
\end{figure}

\textbf{Scope and Organization.} As illustrated in Fig.~\ref{fig:overview}, this survey provides a structured overview of differential equations as a foundational principle for understanding, analyzing, and improving DNNs. We focus on three guiding questions:
\begin{enumerate}
    \item[Q1:] How can differential equations offer a principled understanding of DNN architectures?
    \item[Q2:] How can tools from differential equations be used to improve DNN performance in a principled way?
    \item[Q3:] What real-world applications benefit from grounding DNNs in differential equations?
\end{enumerate}

For the first two questions, we adopt a two-fold perspective: \textit{(i)} \emph{Model-level}, which considers the entire DNN as a differential equation; and (ii) \emph{Layer-level}, which models individual components, including convolutional, recurrent, attention, or SSM layers,  as differential equations. By structuring our discussion around these three core questions from both levels of abstraction, we aim to illuminate emerging methodologies, uncover fundamental insights, and identify open challenges. We hope that this survey will serve both as a foundational reference and as a source of inspiration for future research at the intersection of deep learning and differential equations. 
For the third question, we review a broad spectrum of real-world applications that benefit from the differential equation viewpoint, including time series, generative AI, and large language models, illustrating how differential equation principles can enhance modeling capabilities across diverse domains.

The remainder of this survey is structured as follows. 
Sec.~\ref{sec:related-survey} reviews prior surveys that explore the intersection between differential equations and DNNs, and clarifies the distinctions between our work and existing literature. 
Sec.~\ref{sec:arch-design} discusses how differential equations provide a principled foundation for understanding and designing modern DNN architectures.  
Sec.~\ref{sec:improving} examines how tools and insights from differential equations can be leveraged to improve the performance, robustness, and efficiency of DNNs.  
Sec.~\ref{sec:application} surveys practical applications across various domains that benefit from the differential equation perspective. 
Sec.~\ref{sec:discussion} outlines key challenges and promising future directions in this emerging research area. 
Finally, Sec.~\ref{sec:conclusion} concludes the survey.

\textbf{Notations.} To ensure clarity across the diverse fields addressed in this paper, we establish the following unified notations. Scalars are denoted by regular (non-bold) letters, e.g., $x$ or $X$. Vectors are denoted by lowercase boldface letters, e.g., $\xB$. Matrices are denoted by uppercase boldface letters, e.g., $\XB$.  Correspondingly, functions with scalar, vector, and matrix outputs are denoted using matching letters: $f(\cdot)$, $\fB(\cdot)$, and $\FB(\cdot)$, respectively. Sets are denoted by calligraphic symbols, e.g., $\calA$. The abbreviations used in the survey are summarized in Tab. \ref{tab:abbrv} in Appendix \ref{app:abbrv}.

\section{Distinction from Previous Surveys}\label{sec:related-survey}

Several prior surveys have examined the intersection of differential equations and DNNs, as summarized in Table~\ref{tab:related-surveys}. While these works offer valuable overviews by reviewing model formulations, theoretical properties, or application-specific developments, most are limited in scope. Many focus exclusively on either the \emph{model-level} perspective (e.g., Neural ODEs and their variants) or the \emph{layer-level} perspective (e.g., Deep State Space Models). Furthermore, several surveys concentrate on specific architectural classes or mathematical tools, lacking a unifying perspective that connects model design, theoretical analysis, and performance enhancement.

In contrast, this survey integrates both the \emph{model-level} viewpoint, which interprets the entire DNN as a dynamical system, and the \emph{layer-level} viewpoint, which embeds differential equation structures into individual network components, as illustrated in Fig.~\ref{fig:overview}. Additionally, we adopt a \emph{user-guided} and \emph{problem-driven} approach. Rather than organizing the discussion around specific model classes or techniques, we structure the survey around three fundamental questions that target core challenges for DNNs. This dual-perspective and user-centered framework allows us to unify a broad range of theoretical insights and methodological innovations into a coherent and structured narrative. By bridging foundational mathematical theory with practical concerns, our goal is to provide a comprehensive and actionable roadmap for researchers and practitioners working at the interface of DNNs and differential equations.

\begin{remark}
While this survey focuses on the use of differential equations as a foundational framework for understanding and improving DNNs, a parallel line of research explores the use of DNNs as surrogate solvers for high-dimensional partial differential equations (PDEs). This alternative perspective, exemplified by physics-informed neural networks (PINNs)~\cite{Raissi2017zsi} and operator learning frameworks~\cite{liFourierNeuralOperator2020, luLearningnonlinearoperators2021a}, aims to overcome the computational bottlenecks inherent in traditional PDE solvers. Although highly relevant to the broader intersection of machine learning and differential equations, this topic lies beyond the primary scope of our survey. We refer interested readers to recent overviews and benchmarks~\cite{huangPartialDifferentialEquations2022, takamotoPDEBenchExtensiveBenchmark2022, nganyutanyuDeeplearningmethods2023, kovachkiOperatorLearningAlgorithms2024} for a comprehensive treatment of DNN-based PDE solvers and operator learning methods.
\end{remark}

\begin{table}[htbp]
    \centering
    \caption{Existing surveys on the differential equation foundations of DNNs}
    \label{tab:related-surveys}
    \renewcommand{\arraystretch}{1.3}
    \begin{tabularx}{\linewidth}{p{0.9cm}|c|X}
        \toprule
        \textbf{Scope} & \textbf{Paper} & \textbf{Summary} \\ 
        \midrule
        \multirow{4}{*}{\raisebox{-4.5\totalheight}{%
          \shortstack{Model\\Level}
        }}
        & \cite{ChenReviewOrdinaryDifferential2019} & This survey paper explores ODE-based neural architecture design via discretization, along with continuous modeling and optimal control-inspired optimization. \\
        & \cite{ZhangDynamicalPerspectiveMachine2020} & This work reviews machine learning from a dynamical systems perspective, covering gradient-based and non-gradient optimization methods, continuous-time formulations. \\  
        & \cite{kidgerNeuralDifferentialEquations2022} & This paper systematically reviews neural differential equations (ODEs, CDEs, SDEs, RDEs) and their underlying techniques. \\  
        & \cite{FazakasExploringIntegrationDifferential2024} & This review examines the mathematical foundations of incorporating ordinary and partial differential equations into DNNs, focusing on Neural ODEs and PINNs. \\
        & \cite{ohComprehensiveReviewNeural2025} & This survey provides a comprehensive review of neural differential equations, with a focus on time series analysis. \\
        & \cite{liu2025deep} & This work discusses how different DNN architectures can be viewed the deterministic or stochastic differential equations, and tests them on CIFAR-10 and CIFAR-100. \\
        \midrule
        
        \multirow{4}{*}{\raisebox{-2.3\totalheight}{%
          \shortstack{Layer\\Level}
        }}
        & \cite{AlonsoStateSpaceModels2024} & This paper provides a control-theoretic perspective on SSMs as a foundation for deep learning architectures, particularly in sequence modeling. \\
        & \cite{PatroMamba360SurveyState2024} & This survey summarizes fundamental Deep SSMs by categorizing them into Gating architectures, Structural architectures, and Recurrent architectures. \\
        & \cite{WangStateSpaceModel2024} & This work presents a comprehensive survey of Deep SSMs and their variants, and summarizes their applications in different fields. \\
        & \cite{somvanshi2025survey} & This survey reviews the development of Deep SSMs, provides a comparative analysis of Deep SSMs and Transformers, and discusses their practical applications and future research directions. \\
        \bottomrule
    \end{tabularx}
\end{table}

\section{Understanding DNNs as Differential Equations}\label{sec:arch-design}

In this section, we address the \textbf{first} key question: \emph{How can differential equations offer a principled understanding of DNN architectures?} To explore this question, we review a broad spectrum of works that leverage differential equations for constructing DNNs. Guided by the framework illustrated in Fig.~\ref{fig:overview}, our discussion is organized along two complementary levels of abstraction:

\begin{itemize}
    \item[(i)] \textbf{Model-Level Perspective}: The entire DNN is formulated as a continuous or discretized differential equation, typically grounded in numerical methods for solving ODEs or SDEs. This perspective encompasses a variety of architectures, including ResNets, Neural ODEs, and flow-based generative models such as diffusion models and continuous normalizing flows.

    \item[(ii)] \textbf{Layer-Level Perspective}: Differential equation principles are employed in the design of individual layers within the network. This approach underlies the development of Deep SSMs, which generalize and unify key components such as convolution, recurrence, and attention by embedding them within a shared dynamical system framework.
\end{itemize}

Together, these perspectives offer a mathematically grounded foundation for the design of scalable, expressive, and stable DNN architectures. In the remainder of this section, we first examine the model-level perspective, followed by a discussion of the layer-level approach.

\subsection{Model-Level Perspective}\label{sec:model-level}

The \emph{model-level perspective} is motivated by the observation that many DNNs can be interpreted as numerical approximations of continuous-time dynamical systems. A seminal example is the interpretation of ResNets~\cite{he2016deep} as a forward Euler discretization of ODEs~\cite{EProposalMachineLearning2017, haberStableArchitecturesDeep2018}. Mathematically, the update rule of a ResNet layer can be written as:
\begin{equation}\label{eq:ResNet}
    \hB_{k+1} = \hB_k + \fB_{\thB_k}(\hB_k),
\end{equation}
where $\hB_k$ denotes the hidden state at the $k$-th layer, and $\fB_{\thB_k}(\cdot)$ is a transformation function parameterized by $\thB_k$. This update rule corresponds to the forward Euler discretization of the following ODE:
\begin{equation}\label{eq:forward-euler}
    \dot{\hB}(t) = \fB_{\thB(t)}(t, \hB(t)) \quad \Longrightarrow \quad \hB_{k+1} = \hB_k + \Delta t \fB_{\thB_k}(\hB_k),
\end{equation}
when the step size $\Delta t = 1$. Building on this insight, early research explored alternative skip-connection schemes through the lens of ODE discretizations, ultimately leading to the formulation of \emph{Neural ODEs}. This framework was subsequently generalized to include broader classes of differential equations, such as CDEs and SDEs, giving rise to the more general paradigm of \emph{Neural Differential Equations (NDEs)}. These principles have also been effectively applied in generative modeling, including the development of diffusion models and flow models, in which the sampling is implemented by simulating SDEs or ODEs.

\subsubsection{Early Approaches: Skip Connections as Discretization of ODEs}

\begin{figure}
    \centering
    \includegraphics[width=\linewidth]{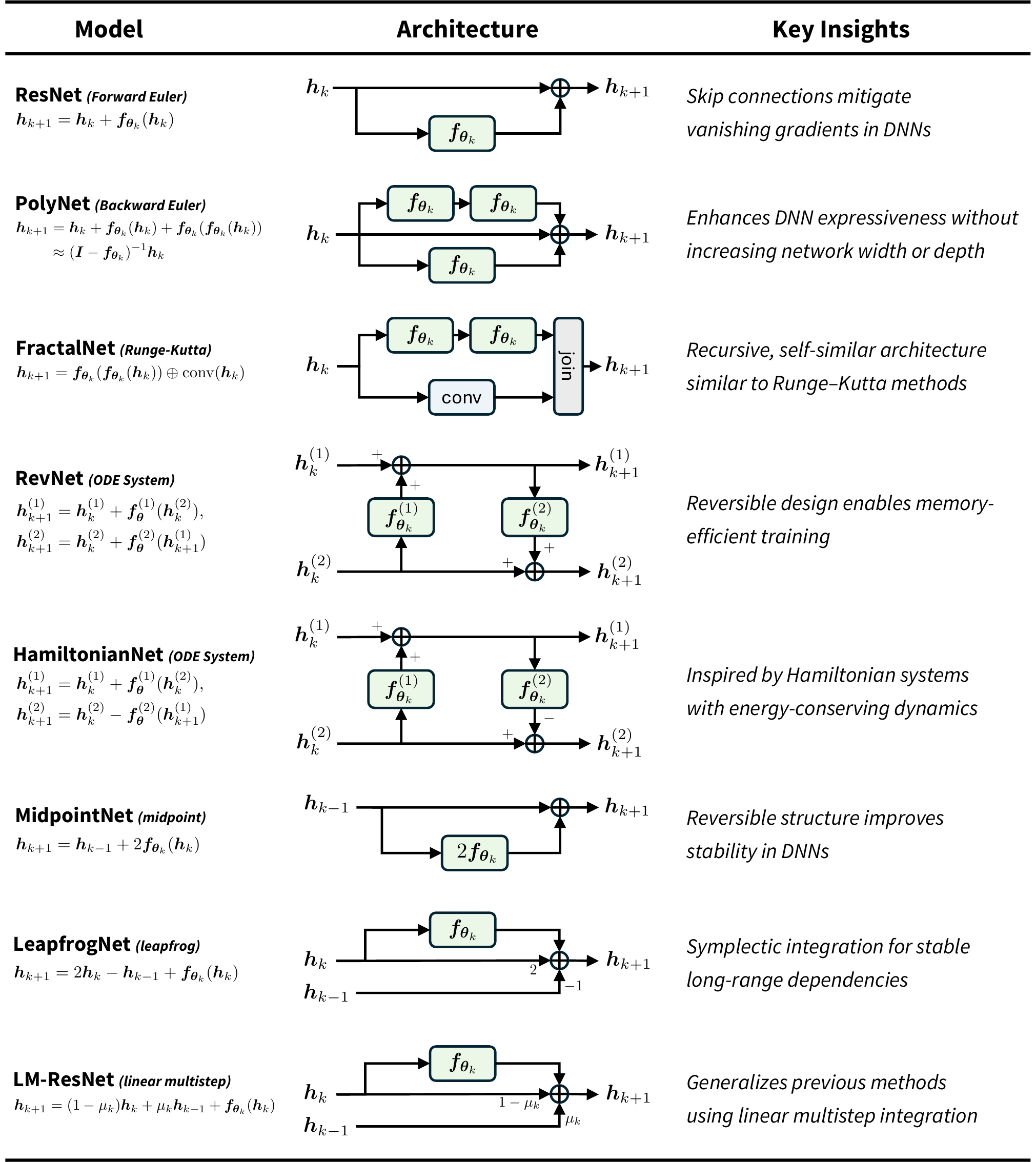}
    \caption{Summary of early approaches that design DNN architectures through the lens of differential equations. A central idea in these works is to interpret various skip-connection schemes as specific discretizations of ODEs, or introduce different ODE systems for special properties. The plus and minus symbols beside the arrows denote additive or subtractive operations, respectively, and the numbers indicate coefficients.}
    \label{fig:resnets}
\end{figure}

Building on the foundational insights of~\cite{EProposalMachineLearning2017, haberStableArchitecturesDeep2018}, a key early research direction involved interpreting skip connections in DNNs as discretizations of ODEs, as summarized in Fig.~\ref{fig:resnets}. Among the first to formalize this connection,~\cite{luFiniteLayerNeural2018a} systematically analyzed how different skip-connection schemes correspond to explicit and implicit ODE solvers. For instance, PolyNet~\cite{ZhangPolyNetPursuitStructural2017} enhanced network expressivity by incorporating polynomial compositions into skip connections. This architecture can be interpreted as an approximation to the \emph{backward Euler} discretization of an ODE, enabling larger step sizes and fewer residual blocks. FractalNet~\cite{LarssonFractalNetUltraDeepNeural2017} adopted recursive expansion rules to generate self-similar structures, which resemble the integration steps of \emph{Runge-Kutta} methods. RevNet~\cite{GomezReversibleResidualNetwork2017}, on the other hand, introduced reversible transformations that allow exact reconstruction of activations, thereby reducing memory usage during training. This design aligns with a \emph{forward Euler} discretization and emphasizes computational efficiency.

Motivated by these discretization-inspired insights, several studies have proposed novel architectures based on alternative numerical schemes. For example,~\cite{changReversibleArchitecturesArbitrarily2018} introduced three reversible models, HamiltonianNet, MidpointNet, and LeapfrogNet, that incorporate dynamical system stability properties into architectural design. LM-ResNet~\cite{luFiniteLayerNeural2018a} unified these variants via a two-step linear multistep (LM) method, yielding a flexible framework with strong empirical performance. Table~\ref{fig:resnets} summarizes the mathematical formulations underlying these discretization-based models. 

While most of these approaches focus on skip connections, similar ideas have also been extended to convolutional networks. For instance, \cite{RuthottoDeepNeuralNetworks2020} established a connection between CNNs and PDEs, showing that forward propagation corresponds to time evolution, and convolutional operations approximate spatial derivatives. They further proposed parabolic, hyperbolic, and second-order hyperbolic CNNs, which align respectively with the discretizations underlying ResNet, HamiltonianNet, and LeapfrogNet in \cite{changReversibleArchitecturesArbitrarily2018}.

These early developments laid the foundation for viewing DNNs from the perspective of differential equations. Building upon these insights, \textit{Neural Differential Equations} emerged as a natural extension --- formulations in which the forward pass of a DNN is defined by directly solving an ODE using numerical solvers. We explore this class of models in detail in the next subsection.

\subsubsection{Neural Differential Equations} 

\begin{figure}[htbp]
    \centering
    \includegraphics[width=0.8\linewidth]{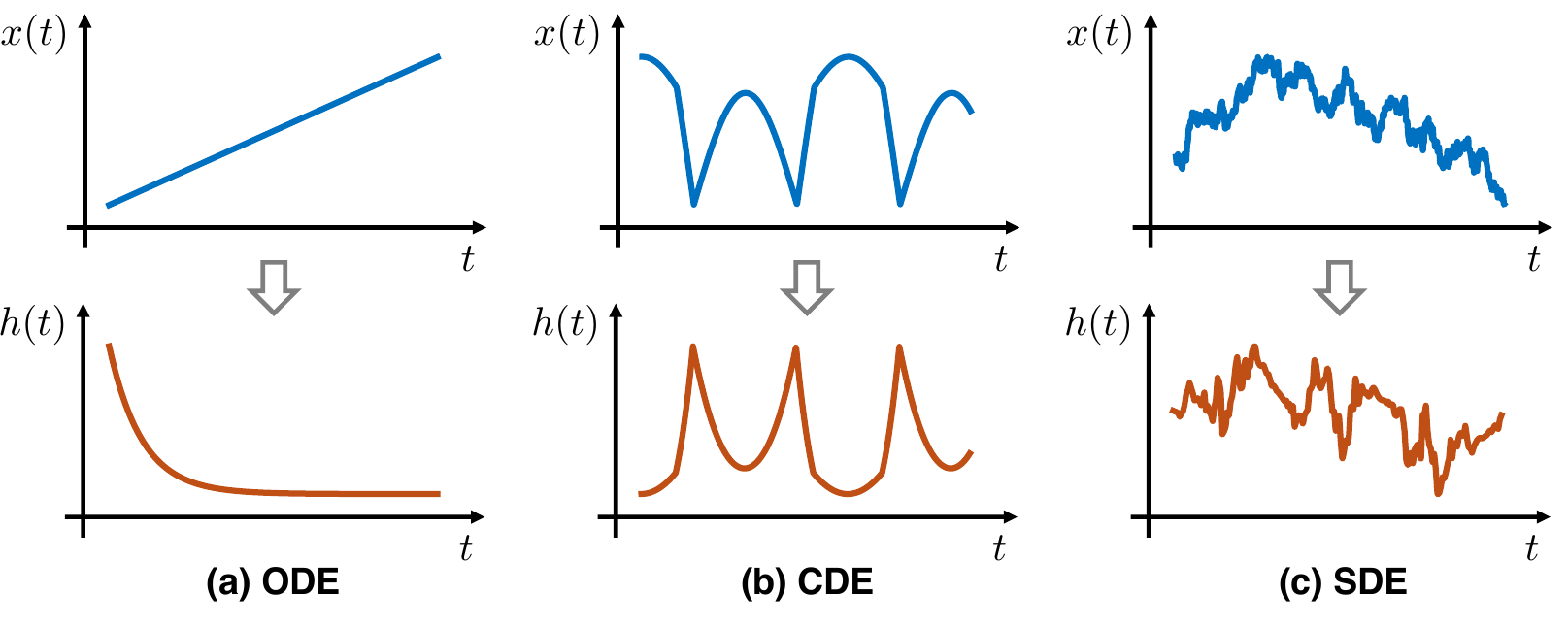}
    \caption{Illustration of three commonly used types of differential equations in NDEs. They share the unified formulation $\dd{\hB} = \FB(\hB)\dd{\xB}$, where the specific type depends on the regularity of the driving signal $\xB(t)$. (a) When $\xB(t) = t$, the system corresponds to an ODE; (b) when $\xB(t)$ has bounded variation, it becomes a CDE; and (c) when $\xB(t)$ is a Brownian motion, it defines an SDE.}
    \label{fig:nde-framework}
\end{figure}

In this subsection, we discuss how various types of differential equations can be unified under the NDE framework. Specifically, we focus on three fundamental classes: ODEs, CDEs, and SDEs, as illustrated in Fig.~\ref{fig:nde-framework}. These formulations share a common structure \cite{friz2020course, lyons2007differential}:
\[
\dd{\hB(t)} = \FB(\hB(t))\dd{\xB},
\]
where the nature of the driving signal $\xB(t)$ determines the specific type of equation. When $\xB(t) = t$, the equation reduces to an ODE. When $\xB(t)$ is a path of bounded variation, the system becomes a CDE. If $\xB(t)$ is a Brownian motion, the system corresponds to an SDE. We now introduce the design principles and modeling considerations behind Neural ODEs, Neural CDEs, and Neural SDEs, respectively.

\textbf{Neural ODEs.}
Building on discretization-based architectures, \emph{Neural ODEs}~\cite{ChenNeuralOrdinaryDifferential2018a} were introduced as continuous-depth models, where the transformation dynamics are governed by an ODE directly parameterized by a DNN:
\begin{equation}\label{eq:neural-ode}
    \dot{\hB}(t) = \fB_{\thB}(t, \hB(t)).
\end{equation}
The hidden state at time $t$ is obtained by solving this ODE:
\begin{equation}
    \hB(t) = \hB(0) + \int_0^t \fB_{\thB}(\tau, \hB(\tau)) \dd{\tau} = \mathrm{ODESolve}(\fB_{\thB}, \hB(0), 0, t),
\end{equation}
where $\mathrm{ODESolve}(\cdot)$ denotes off-the-shelf numerical ODE solvers, such as Euler’s method or higher-order Runge–Kutta schemes \cite{butcher2016numerical}. In contrast to discrete skip-connection architectures in early approaches, which use fixed-step updates, Neural ODEs leverage \textit{adaptive-step solvers} to define the network's forward pass, allowing for continuous-depth computation and improved parameter efficiency. This formulation is particularly appealing for modeling continuous-time processes and handling irregularly sampled data. 

However, backpropagating through the operations of an ODE solver introduces significant memory overhead, as intermediate states must be stored throughout the forward pass. To address this issue, the \emph{adjoint sensitivity method} is commonly employed. This technique introduces an auxiliary ODE that governs the evolution of the adjoint state $\aB(t) = \pdv*{\calL}{\hB(t)}$, where $\calL(\cdot)$ is a scalar loss function. The gradient of $\calL$ with respect to the model parameters $\thB$ can then be computed by solving the following system of ODEs:
\begin{equation}\label{eq:adjoint-method}
    \dot{\aB}(t) = -\aB(t)^\top \pdv{\fB_{\thB}(t, \hB(t))}{\hB(t)}, 
    \quad 
    \pdv{\calL}{\thB} = \int_{t}^{0} \pdv{\fB_{\thB}(t, \hB(t))}{\thB} \dd{t}.
\end{equation}
This method views numerical ODE solvers as black boxes, avoiding storing the full trajectory of $\hB(t)$ and other intermediate variables during the forward pass. Instead, gradients are computed through a third call to the ODE solver in the backward phase, reducing memory usage to $\mathcal{O}(1)$. However, despite its theoretical advantage, the adjoint method faces two major limitations: (i) \emph{Computational inefficiency.} During training, the numerical solver must evaluate the DNN-based vector field $\fB_{\thB}$ numerous times, resulting in considerable overhead. Recent works aim to mitigate this by modifying the solver mechanisms~\cite{kelly2020learning}, regularizing the learned dynamics~\cite{finlay2020train, kelly2020learning}, or introducing solver-free training techniques~\cite{quaglino2019snode, LipmanFlowMatchingGenerative2022, zhao2025accelerating}. (ii) \emph{Numerical instability in the backward pass.} The backward integration of the adjoint system introduces additional discretization errors, which can degrade training stability and model performance. To address this, some methods incorporate \emph{checkpointing strategies} that selectively store intermediate states during the forward pass to improve gradient accuracy~\cite{zhuang2020adaptive}.

Building upon this foundational work, a natural progression has been to generalize the Neural ODE framework to encompass a broader class of differential equations. This has led to the emergence of the unified field of \emph{Neural Differential Equations} (NDEs)~\cite{kidgerNeuralDifferentialEquations2022}, which incorporates not only ODEs but also more expressive formulations that capture external inputs, randomness, and complex temporal dynamics. In what follows, we focus on two representative extensions of this framework: \emph{Neural Controlled Differential Equations} (Neural CDEs), which are well-suited for modeling irregular and input-driven time series, and \emph{Neural Stochastic Differential Equations} (Neural SDEs), which introduce stochasticity into the latent dynamics to capture uncertainty and complex variability.

\begin{table}[tbp]
    \centering
    \caption{Overview of representative Neural Differential Equation (NDE) models. Each formulation extends classical differential equations with neural network parameterizations to model continuous-time dynamics.}
    \label{tab:ndes}
    \resizebox{\textwidth}{!}{%
    \renewcommand{\arraystretch}{1}
    \begin{tabular}{ll}
    \toprule
    \multicolumn{1}{c}{\textbf{Model}} & 
    \multicolumn{1}{c}{\textbf{Formulation and Key Insight}}
    \\
    \midrule
    
    Neural ODE \cite{ChenNeuralOrdinaryDifferential2018a} & 
    \makecell[l]{
        $\dot{\hB}(t)=\fB_{\thB}(t, \hB(t))$ \\ \small
        Models continuous-depth dynamics by taking the deep limit of ResNets
    } 
    \\  \\
    Neural CDE \cite{KidgerNeuralControlledDifferential2020} & 
    \makecell[l]{
        $\dd{\hB(t)} = \FB_{\thB}(\hB(t)) \db{x}(t) \dd{t}$  \\ \small
        Introduces input-dependent dynamics for irregular time series
    } 
    \\ \\
    Neural RDE \cite{MorrillNeuralRoughDifferential2021} &
    \makecell[l]{
        $\dd{\hB(t)} = \FB_{\thB}(\hB(t)) \dd{\mathrm{LogSig}_{r_i, r_{i+1}}(\xB(t)}$ \\ \small
        Incorporates log-signature transforms to model long and rough time series
    }
    \\ \\
    Neural SDE \cite{tzen2019neural} &
    \makecell[l]{
        $\dd{\hB(t)} = \bm{b}_{\thB}(t, \hB(t)) \dd{t} + \bm{\sigma}_{\phiB}(t, \hB(t)) \dd{\wB(t)}$ \\ \small
        Extends Deep Latent Gaussian Models as Neural SDEs in the diffusion limit 
    } 
    \\ \\
    Neural Jump SDE \cite{jia2019neural} &
    \makecell[l]{
        $\dd{\hB(t)} = \bm{f}_{\thB}(t, \hB(t)) \dd{t} + \bm{g}_{\phiB}(t, \hB(t), \kB(t)) \dd{N(t)}$ \\ \small
        Captures hybrid systems with both continuous flows and discrete jumps
    }
    \\ \\
    Latent SDE \cite{li2020scalable} &
    \makecell[l]{
        $\dd{\hb{z}(t)} = \bm{f}_{0, \thB}(t, \zB(t)) \dd{t} + \bm{\sigma}_{\phiB}(t, \zB(t)) \dd{\wB(t)},~ \dd{\zB(t)} = \bm{f}_{1, \thB}(t, \zB(t)) \dd{t} + \bm{\sigma}_{\phiB}(t, \zB(t)) \dd{\wB(t)}$ \\ \small
        Introduces memory-efficient adjoint methods for Neural SDEs in the latent space
    }
    \\ \\
    Neural SDE as GAN \cite{KidgerNeuralSDEsInfiniteDimensional2021} & 
    \makecell[l]{
        $
        \dd{\xB(t)} = \bm{\mu}_{\thB}(t, \xB) \dd{t} + \bm{\sigma}_{\thB}(t, \xB) \circ \dd{\wB(t)}, ~\dd{\hB(t)} = \fB_{\phiB}(t, \hB) \dd{t} + \gB_{\phiB}(t, \hB) \circ \dd{\yB(t)} 
        $ \\ \small
        Interprets SDE calibration as infinite-dimensional GAN training
    } 
    \\  \\
    Stable Neural SDEs \cite{oh2024stable} & 
    \makecell[l]{
        $\dd{\zB(t)} = \bm{\gamma}_{\thB}(t, \zB(t)) \dd{t} + \bm{\sigma}_{\phiB}(t, \zB(t)) \dd{\wB(t)}$ \\ \small
        Proposes Langevin, Linear-Noise, and Geometric SDEs to enhance robustness and stability
    }
    \\
    \bottomrule
    \end{tabular}
    }
\end{table}

\textbf{Neural CDEs.} 
Neural CDEs were introduced by~\cite{KidgerNeuralControlledDifferential2020} to address a key limitation of Neural ODEs: the output trajectory is determined entirely by the initial condition and vector field, without directly incorporating new input over time. Neural CDEs resolve this by treating the input as a continuous-time control signal, enabling dynamic updates as data arrives. Inspired by rough path theory \cite{friz2020course, lyons2007differential}, Neural CDEs are defined as:
\begin{equation}\label{eq:neural-cde}
    \dd{\hB(t)} = \FB_{\thB}(\hB(t)) \dd{\xB(t)} 
    \quad \Longrightarrow \quad 
    \hB(t) = \hB(0) + \int_0^t \FB_{\thB}(\hB(\tau)) \dd{\xB(\tau)},
\end{equation}
where $\FB_{\thB}: \mathbb{R}^{d_h} \rightarrow \mathbb{R}^{d_h \times d_x}$ is a learnable vector field, and $\xB(t)$ is the continuous input path. The integral is taken in the Riemann–Stieltjes sense.
Under mild regularity conditions on $\xB(t)$, the CDE can be rewritten as a standard ODE:
\begin{equation}\label{eq:cde-ode}
    \dd{\hB(t)} = \underbrace{\FB_{\thB}(\hB(t)) \db{x}(t)}_{\hb{f}_{\thB}(t, \xB(t))} \dd{t}
    \quad \Longrightarrow \quad
    \db{h}(t) = \hb{f}_{\thB}(t, \xB(t)),
\end{equation}
where $\db{x}(t)$ denotes the time derivative of the interpolated input path. In this case, it is easy to use existing ODE solvers to solve CDEs. This formulation enables the model to process continuously evolving data in a \emph{time-aware} and \emph{input-dependent} manner, making it particularly suitable for handling irregularly sampled time series. In the original formulation,~\cite{KidgerNeuralControlledDifferential2020} employed natural cubic spline interpolation to construct the input path $\xB(t)$ to approximate $\db{x}(t)$ in Eq~\eqref{eq:cde-ode}, which assumes access to the entire time series in advance, a constraint that limits applicability in real-time settings. To address this,~\cite{morrill2021neural} proposed new interpolation strategies and identified theoretical conditions that support online forecasting with Neural CDEs.

Building on this foundation, several extensions have been introduced. One notable example is the \emph{Neural Rough Differential Equation} (Neural RDE)~\cite{MorrillNeuralRoughDifferential2021}, which generalizes Neural CDEs by incorporating log-signature transforms from rough path theory~\cite{lyons2007differential, friz2020course}, which is a special kind of feature map for rough paths. While maintaining the same structural form, Neural RDEs enhance expressiveness by capturing higher-order statistical information about the input path. This results in improved performance on tasks involving long-range dependencies and irregular temporal patterns.

\textbf{Neural SDEs.}  
SDEs extend ODEs by introducing stochasticity, thereby enabling the modeling of systems influenced by noise or uncertainty. A general form of an SDE is:
\begin{equation}\label{eq:sde}
    \dd{\xB(t)} = \fB(t, \xB(t))\dd{t} + \GB(t, \xB(t))\dd{\wB(t)},
\end{equation}
where $\fB$ is the \emph{drift} term, $\GB$ is the \emph{diffusion} term, and $\wB(t)$ denotes a standard Wiener process (i.e., Brownian motion). To numerically simulate trajectories of~\eqref{eq:sde}, the Euler–Maruyama scheme is typically used:
\begin{equation}\label{eq:euler-maruyama}
    \xB(t + h) = \xB(t) + h\fB(t, \xB(t)) + \sqrt{h}\GB(t, \xB(t))\bm{\varepsilon}, \quad \bm{\varepsilon} \sim \calN(0, \IB),
\end{equation}
where $h > 0$ is the time step size and $\IB$ is the identity matrix of appropriate dimension.

Neural SDEs generalize this formulation by parameterizing both the drift and diffusion terms using DNNs:
\begin{equation}\label{eq:neural-sde}
    \dd{\hB(t)} = \bm{f}_{\thB}(t, \hB(t)) \dd{t} + \bm{G}_{\phiB}(t, \hB(t)) \dd{\wB(t)},
\end{equation}
where $\hB(t)$ denotes the latent (or hidden) state, and the drift $\bm{f}_{\thB}$ and diffusion $\bm{G}_{\phiB}$ are DNNs parameterized by $\thB$ and $\phiB$, respectively.

Early work by~\cite{tzen2019neural} introduced Neural SDEs as a deep generative model, casting them as latent Gaussian processes in the diffusion limit. They also developed a variational inference framework that supports black-box training via automatic differentiation through SDE solvers. To model systems with both continuous dynamics and discrete events,~\cite{jia2019neural} proposed Neural Jump SDEs, which integrate jump processes into SDEs for modeling temporal point processes using piecewise-continuous latent trajectories. To address computational limitations,~\cite{li2020scalable} developed a memory-efficient version of the adjoint sensitivity method tailored for Neural SDEs, and introduced \emph{Latent SDEs} as a generalization of Latent ODEs~\cite{rubanova2019latent}. In a different direction,~\cite{KidgerNeuralSDEsInfiniteDimensional2021} interpreted Neural SDEs as infinite-dimensional generative adversarial networks (GANs), leveraging the insight that traditional SDE fitting can be viewed as a special case of Wasserstein GANs~\cite{arjovsky2017wasserstein}. More recently,~\cite{oh2024stable} highlighted the importance of careful design in both the drift and diffusion networks to ensure model stability. They proposed three new architectures (Langevin-type SDEs, Linear Noise SDEs, and Geometric SDEs) that incorporate structural priors to guarantee robustness and convergence. These stabilized Neural SDEs demonstrate strong empirical performance across a variety of time series tasks, including medical data analysis, financial forecasting, air quality modeling, and motion capture dynamics.

\begin{remark}
    Following the development of NDEs, a number of works have explored hybrid architectures that combine NDEs with conventional neural components to enhance modeling capacity and expressivity. Representative examples include combining Neural ODEs with recurrent neural networks (RNNs) and variational autoencoders (VAEs)~\cite{rubanova2019latent, debrouwerGRUODEBayesContinuousModeling2019, yildiz2019ode2vae, lechner2020learning}, as well as integrating Neural CDEs into Transformer-based models~\cite{chen2023contiformer, moreno2024rough}. Although these models are closely related to the NDE framework, they do not aim to represent the entire neural network as a differential equation, but rather incorporate differential components within larger hybrid systems. As such, they fall outside the primary scope of this survey. For discussions of these hybrid architectures and their applications, we refer interested readers to~\cite{kidgerNeuralDifferentialEquations2022, ohComprehensiveReviewNeural2025}.
\end{remark}

\subsubsection{Flow-based Generative Models}\label{subsec:generative}

Motivated by the framework of NDE, recent research has demonstrated that cutting-edge deep generative architectures, particularly flow-based and diffusion-based models \cite{ho2020denoising, LipmanFlowMatchingGenerative2022}, can be systematically interpreted and designed from lens of ODEs or SDEs \cite{Song:2020hus, albergo2023stochastic, flowsanddiffusions2025}.

To formalize the underlying principles, consider the generic paradigm in deep generative models, where a sample from a simple base distribution is iteratively transformed into a sample from the data distribution:
\[
    \xB_0 \sim p_\mathrm{init}, \quad \xB_k = \fB(\xB_{k-1}), \quad k = 1, \dots, K \quad\Longrightarrow\quad \xB_K \sim p_\mathrm{data} 
\]
where $p_\mathrm{init}$ is a tractable, analytic distribution (typically standard Gaussian distribution) and $p_\mathrm{data}$ denotes the desired data distribution (e.g., images of cats).

\begin{figure}[htbp]
    \centering
    \includegraphics[width=\linewidth]{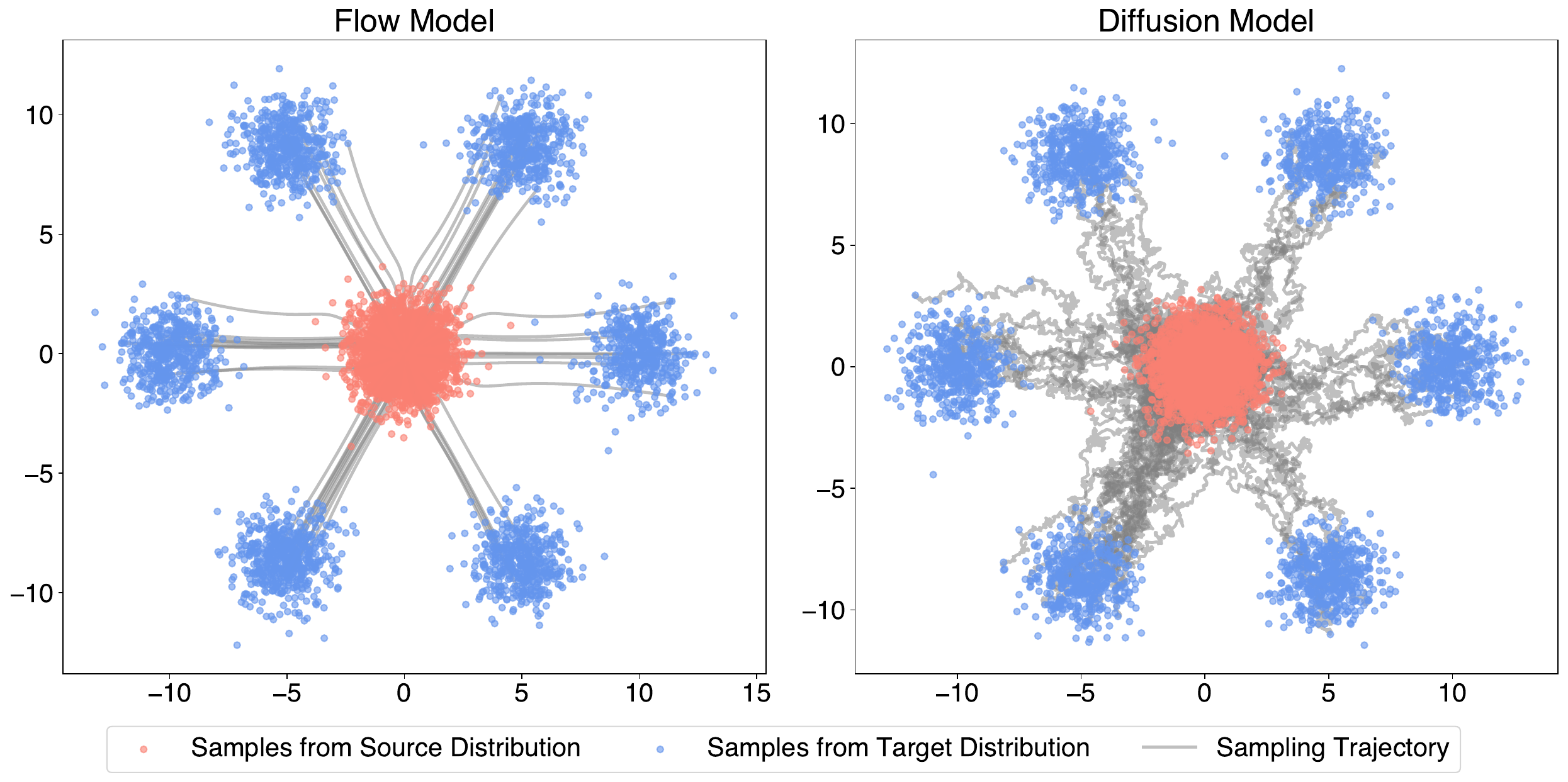}
    \caption{Comparison between flow-based and diffusion-based generative models. Flow models generate samples by simulating ODEs, resulting in smooth and deterministic sampling trajectories. In contrast, diffusion models rely on SDEs, producing stochastic and inherently rough trajectories due to the presence of noise during the sampling process.}
    \label{fig:flow_vs_diffusion}
\end{figure}

Flow-based generative models extend this idea by replacing the discrete transformations 
$\fB(\cdot)$ with continuous-time dynamics governed by a parameterized vector field. Two canonical formulations emerge:
\begin{align}
    &\xB_0 \sim p_\mathrm{init}, \quad \dd{\xB}_t = \uB_t^{\thB}(\xB_t) \dd{t} 
    &\Longrightarrow\quad \xB_1 \sim p_\mathrm{data}, \tag{ODE-based Flow Model} \\
    &\xB_0 \sim p_\mathrm{init}, \quad \dd{\xB}_t = \uB_t^{\thB}(\xB_t) \dd{t} + \sigma_t \dd{\wB}_t 
    &\Longrightarrow\quad \xB_1 \sim p_\mathrm{data}. \tag{SDE-based Diffusion Model}
\end{align}
Here, $\uB_t^{\thB}(\cdot)$ is a time-dependent vector field modeled by a neural network, $\sigma_t$ is the diffusion coefficient (possibly time-varying), and $(\wB_t)_{0 \le t \le 1}$ represents the Brownian motion. By simulating the ODE or SDE, one obtains samples from the target distribution $p_\mathrm{data}$. Illustrative examples of sampling trajectories for flow-based and diffusion-based models are shown in Fig.~\ref{fig:flow_vs_diffusion}.

To train such models efficiently without relying on expensive trajectory simulations, recent methods adopt the frameworks of Conditional Flow Matching (CFM) and Conditional Score Matching (CSM). These methods define a continuous interpolation path 
$(p_t)_{0 \le t \le 1}$ connecting $p_0 = p_\mathrm{init}$ and $p_1 = p_\mathrm{data}$, and formulate training objectives using the \emph{continuity equation} (for deterministic dynamics) or the \emph{Fokker–Planck equation} (for stochastic dynamics). In CFM, the model learns to approximate the conditional velocity field $\uB(\xB|\zB)$ (where $\zB$ is a sample from $p_\mathrm{data}$) along this path, while in CSM, it learns the conditional score function (i.e., the gradient of the log-density $\grad \log p_t(\xB)$). For detailed formulations and practical implementations, we refer readers to \cite{flowsanddiffusions2025, LipmanFlowMatchingGuide2024, albergo2023stochastic}.

We now briefly review the historical development of flow-based generative models. 
\emph{Diffusion models} were first introduced in \cite{sohl2015deep} and significantly advanced by the Denoising Diffusion Probabilistic Models (DDPM) framework in \cite{ho2020denoising}. These models rely on training a neural network to reverse a forward diffusion process that gradually corrupts data into a simple prior (typically Gaussian noise). While early approaches formulated training via variational inference, later work such as \cite{Song:2020hus} reinterpreted these models as score-based generative models. This perspective frames the generative process as a reverse-time SDE driven by a learned score function that approximates the gradient of the log-density of intermediate noisy distributions. The generative trajectory thus inverts the forward-time SDE that originally perturbed the data.

\emph{Flow-based models}, particularly Continuous Normalizing Flows (CNFs), were introduced in the context of Neural ODEs \cite{ChenNeuralOrdinaryDifferential2018a}, which is a special case of normalizing flows \cite{papamakarios2021normalizing}. CNFs model the transformation between a base and target distribution via a continuous-time, invertible ODE, with exact likelihood computation enabled through the instantaneous change of variables formula. However, training CNFs is often computationally intensive due to the need for repeated numerical integration. To mitigate this, \emph{flow matching} \cite{LipmanFlowMatchingGenerative2022, LipmanFlowMatchingGuide2024} techniques were proposed as a simulation-free alternative. These methods avoid ODE solvers by directly learning the time-dependent vector field via supervised regression on samples drawn from known interpolation paths.

\emph{Stochastic interpolants} \cite{albergo2022building, albergo2023stochastic} have recently emerged as a unifying framework that generalizes both CNFs and diffusion models. These methods construct interpolation processes -- either deterministic (ODE-based) or stochastic (SDE-based) -- that smoothly connect arbitrary pairs of distributions through differential equations. This general formulation encompasses score-based diffusion models and CNFs as special cases, offering a principled and flexible foundation for modern generative modeling.

\begin{table}[tbp]
    \centering
    \caption{Overview of differential equation-based generative models. Each model defines a continuous-time stochastic or deterministic process for data generation, typically trained via score matching or flow-based objectives.}
    \label{tab:generative-models}
    \resizebox{\textwidth}{!}{%
    \renewcommand{\arraystretch}{1}
    \begin{tabular}{ll}
    \toprule
    \multicolumn{1}{c}{\textbf{Model}} & 
    \multicolumn{1}{c}{\textbf{Formulation and Key Insight}}
    \\
    \midrule

    Flow Model \cite{ChenNeuralOrdinaryDifferential2018a, LipmanFlowMatchingGenerative2022} & 
    \makecell[l]{
        $\dd{\xB}_t = \uB_t^{\thB}(\xB_t) \dd{t}$  \\ \small
        Trained via flow matching or maximum likelihood; generation performed by simulating ODEs
    } 
    \\ \\
    
    Diffusion Model \cite{ho2020denoising, Song:2020hus} & 
    \makecell[l]{
        $\dd{\xB}_t = \uB_t^{\thB}(\xB_t) \dd{t} + \sigma_t \dd{\wB}_t$ \\ \small
        Trained via score matching; generation achieved by solving SDEs
    } 
    \\ 
    \bottomrule
    \end{tabular}
    }
\end{table}

\subsection{Layer-Level Perspective}

Beyond the model-level perspective, a complementary line of research has emerged that focuses on embedding differential equation structures at the \emph{layer level}. A prominent approach in this direction is the development of \emph{Deep State Space Models} (Deep SSMs), which integrate the classical state space formulation from control theory \cite{KalmanNewApproachLinear1960} into DNN architectures.

State Space Models (SSMs)~\cite{KalmanNewApproachLinear1960} constitute a special class of ODEs that integrate external input signals, making them particularly well-suited for modeling dynamical systems under control. A general SSM is defined as:
\begin{equation}\label{eq:general-SSM}
    \dot{\xB}(t) = \fB(t, \xB(t), \uB(t)), \quad \yB(t) = \gB(t, \xB(t), \uB(t)),
\end{equation}
where $\xB: [0, T] \to \bbR^{d_x}$ represents the system state, $\uB: [0, T] \to \bbR^{d_u}$ denotes the control input, and $\yB: [0, T] \to \bbR^{d_y}$ is the output. The first equation governs the evolution of the internal state, while the second defines the observable output as a function of both the state and input. A widely studied subclass of SSMs is the \emph{linear} SSM. In the time-varying case, it takes the form:
\begin{equation}\label{eq:ltv-ssm}
    \begin{aligned}
        \dot{\xB}(t) &= \AB(t) \xB(t) + \BB(t) \uB(t), \\
        \yB(t) &= \CB(t) \xB(t) + \DB(t) \uB(t),
    \end{aligned}
\end{equation}
where the system matrices $\AB(t)$, $\BB(t)$, $\CB(t)$, and $\DB(t)$ may vary with time $t$. When the system matrices are constant, the model simplifies to a linear time-invariant (LTI) system:
\begin{equation}\label{eq:lti-ssm}
    \begin{aligned}
        \dot{\xB}(t) &= \AB \xB(t) + \BB \uB(t), \\
        \yB(t) &= \CB \xB(t) + \DB \uB(t).
    \end{aligned}
\end{equation}

The recent surge of interest in integrating SSMs into DNN layers is driven by two key lines of research. The first is the development of the \emph{High-order Polynomial Projection Operator (HiPPO)} framework~\cite{GuHiPPORecurrentMemory2020}, which targets long-range sequence modeling. The core idea is to project a continuous-time signal onto a basis of orthogonal polynomials and use the resulting coefficients to represent the sequence. Notably, this work demonstrated that the online update of these coefficients can be formulated as a linear SSM, thereby revealing the potential of SSMs for efficient and scalable long-sequence modeling. The second line of work, proposed in~\cite{GuCombiningRecurrentConvolutional2021a}, establishes that two fundamental DNN operations, convolution and recurrence, can both be interpreted within the SSM framework. Building on this, \cite{DaoTransformersAreSSMs2024} further showed that SSMs can also be connected to attention mechanisms in Transformers~\cite{vaswaniAttentionAllYou2017, KatharopoulosTransformersAreRNNs2020}, by expressing attention using structured matrices. Together, these directions highlight the versatility and unifying potential of differential equation-inspired approaches at the \emph{layer level}.

In the remainder of this section, we first discuss the general formulation of SSM-based layers and their connections to foundational DNN layers such as CNNs, RNNs, and attention. We then review the recent advancements in the development of high-performance SSM architectures.

\subsubsection{SSM-based Layers}

\textbf{General Form of SSM Layers.}
SSM layers were first introduced by \cite{GuCombiningRecurrentConvolutional2021a} as \emph{Linear State Space Layers} (LSSLs). In their simplest form, SSM layers adopt the structure of an LTI system in continuous time, as given by Eq.~\eqref{eq:lti-ssm}:
\begin{equation}\label{eq:ssm-layer} 
    \begin{aligned} 
        \dot{\hB}(t) &= \AB \hB(t) + \BB x(t), \\ 
        y(t) &= \CB \hB(t),
    \end{aligned} 
\end{equation} 
where $x(t) \in \bbR$ and $y(t) \in \bbR$ are scalar-valued input and output signals, and $\hB(t) \in \bbR^{d_h}$ denotes the hidden state. In practice, SSM layers are typically implemented in discrete time, as input sequences are usually sampled discretely. Discretizing Eq.~\eqref{eq:ssm-layer} yields:
\begin{equation}\label{eq:discrete-ssm-layer} 
    \begin{aligned}
        \hB_k &= \bb{A} \hB_{k-1} + \bb{B} x_k, \\
        y_k &= \bb{C} \hB_k,
    \end{aligned}
\end{equation}
where $\bb{A}$, $\bb{B}$ and $\bb{C}$ are discretized versions of $\AB$ and $\BB$ under a time step $\Delta$. Several discretization schemes are commonly used to derive these discrete-time SSMs, including:
\begin{itemize}
    \item \textbf{Forward Euler}: $\bb{A} = \IB + \Delta \cdot \AB$, \quad $\bb{B} = \Delta \cdot \BB$, \quad $\bb{C} = \CB$.
    \item \textbf{Backward Euler}: $\bb{A} = (\IB - \Delta \cdot \AB)^{-1}$, \quad $\bb{B} = (\IB - \Delta \cdot \AB)^{-1} \Delta \cdot \BB$, \quad  $\bb{C} = \CB$.
    \item \textbf{Bilinear (Tustin)}: $\bb{A} = (\IB - \frac{\Delta}{2} \cdot \AB)^{-1} (\IB + \frac{\Delta}{2} \cdot \AB)$, \quad $\bb{B} = (\IB - \frac{\Delta}{2} \cdot \AB)^{-1} \Delta \cdot \BB$, \quad $\bb{C} = \CB$.
    \item \textbf{Zero-Order Hold (ZOH)}: $\bb{A} = \exp(\AB \cdot \Delta)$, \quad  $\bb{B} = \AB^{-1} (\exp(\Delta \cdot \AB) - \IB) \BB$, \quad $\bb{C} = \CB$.
\end{itemize}


\begin{remark}
    The main distinction between Eq.~\eqref{eq:lti-ssm} and Eq.~\eqref{eq:ssm-layer} lies in the fact that SSM layers typically operate on scalar inputs and outputs, i.e., $x(t) \in \mathbb{R}$ and $y(t) \in \mathbb{R}$. Consequently, the system matrices are shaped as $\AB \in \mathbb{R}^{d_h \times d_h}$, $\BB \in \mathbb{R}^{d_h \times 1}$, and $\CB \in \mathbb{R}^{1 \times d_h}$. When the actual input dimension is $d_u$, the layer comprises $d_u$ independent SSMs (i.e., one for each input channel) resulting in no direct interaction between channels. To overcome this, recent SSM-based architectures~\cite{GuEfficientlyModelingLong2022, GuMambaLinearTimeSequence2023} typically append a pointwise feedforward layer to merge the outputs across channels. Furthermore, $\DB$ is commonly omitted in Deep SSM formulations, as it plays a role analogous to the skip connections in ResNets~\cite{he2016deep}. For simplicity and consistency, we adopt this convention throughout our discussion and exclude the $\DB$ term from the SSM layer formulation.
\end{remark}

\textbf{Connections between fundamental DNN mechanisms and Deep SSMs.}  
A particularly appealing property of Deep SSMs is their ability to serve as a \emph{general-purpose sequence modeling framework}, owing to their connections with several widely used mechanisms in DNNs, including convolution, recurrence, and attention.

\begin{itemize}
    \item \textbf{Convolution:} LTI SSMs can be equivalently represented as convolutions, where the kernel is given by the system's impulse response. This allows them to act as global convolutions over the entire input sequence.
    
    \item \textbf{Recurrence:} On the one hand, discretized linear SSMs naturally represent linear recurrence relations. On the other hand, general RNNs can also be viewed as nonlinear SSMs, making recurrence a special case within the broader SSM framework.
    
    \item \textbf{Attention:} Recent work has established explicit connections between Deep SSMs and linear attention mechanisms by representing both using structured matrices. This connection suggests that SSMs are equivalent to attention under appropriate conditions.
\end{itemize}

We elaborate on each of these connections in the following subsections.

\emph{i) Convolution and SSMs.} 
LTI SSMs are naturally connected to convolution, as their outputs can be computed by convolving the system’s impulse response with the input signal. Specifically, for continuous-time LTI SSMs in Eq.~\eqref{eq:ssm-layer}, the output can be expressed as the following convolution \cite{gu2023modeling}:
\begin{equation}
    y(t) = (K * x)(t) = \int_0^{\infty} K(\tau) \cdot x(t - \tau) \dd{\tau}, \qquad \text{where} \quad K(t) = \CB \exp(t \AB) \BB.
\end{equation}
Here, the function $K(t)$ is referred to as the \emph{impulse response}, which characterizes the output of the system when the input $x(t)$ is the Dirac delta function $\delta(t)$. This impulse response can be derived analytically via the inverse Laplace transform of the system's transfer function $G(s) = \CB(s\IB - \AB)^{-1}\BB$.

Similarly, discretized SSM in Eq.~\eqref{eq:discrete-ssm-layer} can also be expressed in the convolution form. Assuming the initial hidden state $\hB_{-1} = \bm{0}$, the output $y_k$ can be obtained by unrolling the recurrence defined by the two equations in Eq.~\eqref{eq:discrete-ssm-layer}:
\begin{equation}
    y_k = \sum_{i=0}^{k} \bb{C} \bb{A}^{i} \bb{B} x_{k-i} = \bb{C} \bb{A}^k \bb{B} x_0 + \bb{C} \bb{A}^{k-1} \bb{B} x_1 + \cdots \bb{C} \bb{A}\bb{B} x_{k-1} + \bb{C} \bb{B} x_k
\end{equation}
This recursion naturally leads to a discrete convolution:
\begin{equation}\label{eq:conv-ssm}
    \begin{aligned}
    \mathbf{y} &= \mathbf{x} * \bb{K}, \quad \bb{K} = (\bb{C}\bb{B}, \bb{C} \bb{A}\bb{B}, \dots, \bb{C} \bb{A}^{L-1} \bb{B}, \bb{C} \bb{A}^L \bb{B}),
    \end{aligned}
\end{equation}
where $\mathbf{x} = [x_0, x_1, \dots, x_L]^\top$, $\mathbf{y} = [y_0, y_1, \dots, y_L]^\top$, and $\bb{K}$ denotes the \emph{state space convolution kernel}.

Compared to traditional CNNs, a key distinction is that the state space kernel $\bb{K}$ is \emph{global}~\cite{li2023what}, with its kernel size equal to the length of the input sequence. At the same time, the structured nature of the state space convolution kernel enables it to effectively capture long-term dependencies. In addition, a major advantage of this convolutional perspective lies in enabling the efficient training of Deep SSMs. Unlike the original recurrent formulation in Eq.~\eqref{eq:discrete-ssm-layer}, which requires sequential updates across time steps, the convolutional form allows the entire output sequence $\yB$ to be computed in a single batched operation. This enables highly parallel computation on modern hardware and significantly accelerates the training of Deep LTI SSMs such as S4 \cite{GuEfficientlyModelingLong2022}.

\emph{ii) Recurrence and SSMs.} The discrete-time formulation of LTI SSMs in Eq.~\eqref{eq:discrete-ssm-layer} defines a linear recurrence:
\[
\hB_k = \bb{A} \hB_{k-1} + \bb{B} x_k, \qquad y_k = \CB \hB_k,
\]
which corresponds to a linear RNN without activation functions. In fact, \cite{GuCombiningRecurrentConvolutional2021a} demonstrated that even gating mechanisms in classical RNNs can be viewed a kind of discretization. More generally, nonlinear RNNs, such as gated recurrent units (GRUs)~\cite{cho2014learning}, can be interpreted as discretized nonlinear SSMs. A standard GRU update can be written as:
\begin{equation}
    \begin{aligned} 
        &\iB_j = \mathrm{sigmoid}(\WB_1 \xB_j + \WB_2 \hB_j + \bB_1), \\
        &\rB_j = \mathrm{sigmoid}(\WB_3 \xB_j + \WB_4 \hB_j + \bB_2), \\ 
        &\nB_j = \tanh(\WB_5 \xB_j + \bB_3 + \rB_j \odot (\WB_6 \hB_j + \bB_4)), \\
        &\hB_{j+1} = \nB_j + \iB_j \odot (\hB_j - \nB_j), 
    \end{aligned} 
\end{equation} 
which can be interpreted as a discretized nonlinear SSM. Its continuous-time analogue, known as the GRU-ODE~\cite{debrouwerGRUODEBayesContinuousModeling2019}, takes the form:
\begin{equation}
    \begin{aligned}
        \iB(t) &= \mathrm{sigmoid}(\WB_1 \xB(t) + \WB_2 \hB(t) + \bB_1), \\
        \rB(t) &= \mathrm{sigmoid}(\WB_3 \xB(t) + \WB_4 \hB(t) + \bB_2), \\
        \nB(t) &= \mathrm{tanh}(\WB_5 \xB(t) + \bB_3 + \rB_j \odot (\WB_6 \hB(t) + \bB_4)), \\
        \db{h}(t) &= (1 - \iB(t)) \odot (\nB(t) - \hB(t)).
    \end{aligned}
\end{equation}
This connection implies that nonlinear RNNs fall within a general class of nonlinear SSMs. Recent works have further developed RNN architectures directly inspired by state space principles, including RetNet~\cite{sun2023retentive}, RG-LRU~\cite{de2024griffin}, and RWKV~\cite{peng2023rwkv}.

The primary advantage of the recurrent form lies in its efficiency during inference. Specifically, the convolutional form requires access to the entire sequence of past inputs $\xB$ in order to compute each output, which can lead to substantial computational overhead for long sequences. In the recurrent form, however, the hidden state $\hB$ serves as a compact summary of the past, enabling the model to compute the next output using only the current input and the updated state. 

\emph{iii) Linear attention and SSMs.} \cite{DaoTransformersAreSSMs2024} established a principled connection between linear attention mechanisms and SSMs through the notion of \emph{state space duality} (SSD). Under this framework, both SSM layers and linear attention can be interpreted as sequence-to-sequence transformations of the form:
\begin{equation} 
    \yB = \MB_{\thB} \xB, 
\end{equation} 
where $\MB_{\thB} \in \bbR^{(L+1) \times (L+1)}$ denotes the transformation matrix, and $\thB$ comprises the learnable parameters, originating from either the linear attention mechanism or the SSM layer. The core insight is that the matrix $\MB_{\thB}$ induced by linear attention exhibits a structure closely aligned with that generated by SSM layers. This structural equivalence opens the door for transferring techniques between the two domains. For example, importing architectural innovations and theoretical tools from the Transformer literature to guide the design of SSM-based models. Such a connection has recently led to hybrid designs and new perspectives on efficient sequence modeling.

\subsubsection{Development of SSM Layers}
In this section, we review several key developments in the evolution of Deep SSMs and summarize them in Table~\ref{tab:ssm-summary}. The foundation of modern Deep SSMs originates from the HiPPO framework~\cite{GuHiPPORecurrentMemory2020}, a memory mechanism designed for long-sequence modeling. The core idea is to project input signals onto a series of orthogonal polynomial bases (e.g., Legendre polynomials), and to represent the signal using the coefficients associated with these bases. Crucially, the update rule for these coefficients over time can be formulated as a linear time-invariant (LTI) state space model.

The HiPPO mechanism inspired the development of trainable Deep SSM layers. The initial formulation, known as Linear State Space Layers (LSSL), demonstrated the ability to unify multiple fundamental DNN components, including convolutional, recurrent, and continuous-time models, within a single framework. Moreover, incorporating HiPPO matrices endowed LSSLs with strong capabilities for modeling long-range dependencies. However, LSSLs suffered from significant computational inefficiencies, which limited their practical scalability.

To address this limitation, \cite{GuEfficientlyModelingLong2022} introduced the Structured State Space Sequence Model (S4). S4 exploits the diagonal-plus-low-rank (DPLR) form of HiPPO matrices to enable efficient computation. By leveraging a truncated generating function, S4 reduces the computation of high matrix powers to a matrix inversion problem. Then, using the Woodbury identity and Cauchy kernel properties, it implements the matrix operations efficiently. This results in substantial speedups while preserving modeling power. Nonetheless, the implementation of S4 remains technically intricate due to the complexity of the underlying algorithms.

To simplify the parameterization of SSMs while preserving their modeling power, subsequent works investigated whether the low-rank correction in DPLR could be removed. A key result in this direction is the Diagonal State Space (DSS) model proposed by \cite{gupta2022diagonal}, which demonstrated empirically that performance comparable to S4 can be achieved using only diagonal state matrices. Building on this, \cite{GuParameterizationInitializationDiagonal2022} provided a theoretical justification for DSS, showing that the diagonal restriction of S4 surprisingly recovers the same kernel in the limit of infinite state dimension. This work also introduced S4D, a simplified and more flexible variant that incorporates various initialization schemes from S4.

Parallel to this line of research, several alternative formulations of SSMs have been explored. \cite{hasani2023liquid} combined liquid time-constant (LTC) networks \cite{hasani2021liquid} with S4, resulting in Liquid S4, which enhances the generalization and causal reasoning capabilities of SSM layers. In contrast to earlier SSM variants that are single-input, single-output (SISO), \cite{smith2023simplified} proposed S5, a multi-input, multi-output (MIMO) variant of S4 that supports efficient training via parallel scan and handles inputs sampled at variable rates.

More recently, \cite{GuMambaLinearTimeSequence2023} introduced Mamba, also referred to as Selective SSMs or S6. The core innovation of Mamba is its use of linear time-varying SSMs, where parameters are functions of the input. This design allows Mamba to scale effectively to complex sequence modeling tasks, such as language modeling. Subsequently, \cite{DaoTransformersAreSSMs2024} established a theoretical connection between Mamba and the attention mechanism in Transformers through State Space Duality (SSD). This connection enables the application of system-level optimization techniques (such as tensor parallelism) to Deep SSMs, further improving their training efficiency and scalability. For a more comprehensive overview of recent developments in this area, we refer interested readers to the surveys \cite{AlonsoStateSpaceModels2024, PatroMamba360SurveyState2024, WangStateSpaceModel2024, somvanshi2025survey}.

\begin{table}[tbp]
    \centering
    \caption{Summary of key development of Deep SSMs}
    \label{tab:ssm-summary}
    \renewcommand{\arraystretch}{2.0}
    \begin{tabular}{C{0.1\textwidth}L{0.3\textwidth}L{0.25\textwidth}L{0.25\textwidth}}
    \toprule
    \textbf{Model} & \textbf{Summary} & \textbf{Pros} & \textbf{Cons} \\ \midrule
    LSSL \cite{GuCombiningRecurrentConvolutional2021a} 
    & First to introduce linear SSMs as learnable layers in deep networks. 
    & Unifies the advantages of recurrent, convolutional, and continuous-time models.
    & May suffer from memory inefficiency and numerical instability.
    \\ \hline
    
    S4 \cite{GuEfficientlyModelingLong2022}
    & Introduces HiPPO for long-range sequence modeling and Diagonal Plus Low-Rank (DPLR) parameterization for the state matrix $\AB$. 
    & HiPPO enables effective modeling of long-range dependencies, while DPLR ensures efficient training.
    & Hard to implement; built on LTI SSMs with fixed coefficients; limited in handling complex tasks such as language modeling.
    \\ \midrule

    DSS \cite{gupta2022diagonal}
    & Parameterizes the state matrix $\AB$ as a diagonal matrix without low-rank correction, while matching S4's performance.  
    & Simpler to implement, analyze, and train, with provable expressiveness equivalent to general SSMs.  
    & The diagonal structure limits interactions between state dimensions, potentially reducing modeling flexibility.  
    \\ \midrule

    S4D \cite{GuParameterizationInitializationDiagonal2022}  
    & Incorporates various S4 initialization techniques while maintaining a diagonal state matrix.  
    & Simpler to implement, more efficient, and more flexible than DSS.  
    & May sacrifice some expressiveness compared to full or low-rank parameterizations.  
    \\ \midrule

    Liquid S4 \cite{hasani2023liquid}  
    & Combines liquid time-constant (LTC) networks with S4.  
    & Enhances generalization and causal modeling by integrating liquid networks, achieving improved performance across tasks.  
    & The more complex state matrix increases implementation and training difficulty compared to the original S4.  
    \\ \midrule
    
    S5 \cite{smith2023simplified}
    & Introduces multi-input, multi-output (MIMO) SSMs and parallel scan algorithms for efficient training. 
    & Enables modeling of sequences with variable sampling rates while retaining S4-level efficiency.  
    & Still based on LTI SSMs; limited in handling complex sequence modeling tasks.  
    \\ \midrule


    
    Mamba \cite{GuMambaLinearTimeSequence2023}
    & Proposes input-dependent, linear time-varying SSMs with a hardware-efficient parallel scan algorithm for fast recurrent-mode training.
    & Scales to complex sequence tasks across modalities, including language, audio, and genomics.
    & Ecosystem and community support are still developing.
    \\ \midrule

    Mamba-2 \cite{DaoTransformersAreSSMs2024}
    & Introduces state space duality (SSD) to establish a theoretical connection between attention mechanisms and SSMs; leverages block decomposition and system-level optimizations for faster training.
    & Bridges Deep SSMs and Transformers, enabling Mamba to incorporate efficient training techniques from the Transformer literature.
    & Hard to implement    \\
    \bottomrule
    \end{tabular}
\end{table}

\section{Advancing DNNs using Differential Equations}\label{sec:improving}

In this section, we aim to answer the \textbf{second key question}: \emph{How can tools from differential equations be used to improve DNN performance in a principled way?} As a well-established field in applied mathematics, differential equations offer a rich set of analytical tools that can be used to study the behavior of dynamical systems, such as stability theory, forward invariance, and controllability. By interpreting DNNs as instances of differential equations, we can leverage these tools to analyze various properties of DNNs and, in turn, develop methods to improve their performance from a principled standpoint. Following the structure of the previous section, our discussion is organized at two levels: the \emph{model level} and the \emph{layer level}.

\subsection{Advancing DNNs at Model Level}

\textbf{Stability analysis.} 
In control theory, stability is a fundamental property, as it determines whether a system can converge to a desired state despite perturbations. When DNNs are interpreted as discretized differential equations, classical stability tools can be employed to analyze training dynamics and enhance robustness, especially against adversarial attacks.

For example, stability analysis has been used to address vanishing and exploding gradient issues in DNNs. In \cite{haberStableArchitecturesDeep2018}, the authors analyzed the forward propagation of residual networks by relating them to \textit{linear ODEs}. They showed that a necessary condition for stability and well-posedness is that the real parts of the eigenvalues of the weight matrices remain close to zero:
\begin{equation}\label{eq:lin-stable-condition}
    \Re[\lambda_i(\WB_k)] \approx 0,
\end{equation}
where $\lambda_i(\WB_k)$ denotes the $i$-th eigenvalue of the weight matrix $\WB_k$ at layer $k$. The underlying intuition is as follows: if $\Re[\lambda_i(\WB_k)] > 0$, the system exhibits unstable dynamics, leading to exploding gradients; if $\Re[\lambda_i(\WB_k)] \ll 0$, the dynamics become overly contractive, causing vanishing gradients. To address this, the authors proposed two stability-enhancing architectures: (i) DNNs with antisymmetric weight matrices, and (ii) Hamiltonian-inspired DNNs discretized via the Verlet integration method. Both architectures maintained competitive accuracy while substantially improving training stability.

Stability analysis has also been employed to improve the robustness of DNNs. The core idea is that if the underlying dynamical system is stable, it can converge to the correct output despite perturbations in the input or initialization. A central tool in this context is the \emph{Lyapunov function}~\cite{haddad2008nonlinear}, which can be interpreted as a generalized energy or potential function. For a system to be stable, the Lyapunov function should decrease monotonically along its trajectories and reach a minimum at the equilibrium point. 
Formally, consider an ODE system of the form $\dot{\xB} = \fB(\xB)$, where $\xB \in \calX \subseteq \bbR^d$. If there exists a function $V: \bbR^d \to \bbR$ and a point $\xB^* \in \calX$ such that: 
\emph{i)} $V(\xB^*) = 0$,  
\emph{ii)} $V(\xB) > 0$ for all $\xB \in \calX \setminus \{\xB^*\}$, and  
\emph{iii)} $\dv*{V}{t} \leq 0$ for all $\xB \in \calX$,  
then the system is said to be stable in the Lyapunov sense at the equilibrium point $\xB^*$.

Several recent works have leveraged Lyapunov theory to enhance DNN robustness. For instance, \cite{RodriguezLyaNetLyapunovFramework2022} introduced a Lyapunov loss for Neural ODEs by reinterpreting standard loss functions (e.g., cross-entropy) as Lyapunov functions, which are positive-definite functions that are minimized along the system's evolution. This formulation encourages the DNN dynamics to converge quickly to correct predictions while providing improved adversarial robustness. Similarly, \cite{KangStableNeuralODE2021a} linearized Neural ODEs and applied Lyapunov theory to the linearized system to infer global stability. Extending this idea, \cite{chuLyapunovstabledeepequilibrium2024} applied Lyapunov-based analysis to Deep Equilibrium Models (DEQs)~\cite{bai2019deep}, which aim to directly learn fixed points of the DNN's forward dynamics.

\textbf{Forward invariance.}
In the context of differential equations, a set $\mathcal{C}$ is said to be \emph{forward invariant} if, once the trajectory of a dynamical system enters $\mathcal{C}$, it remains in $\mathcal{C}$ for all future time. Formally, for an ODE $\dot{\xB} = \fB(\xB)$, forward invariance is defined as:
\begin{equation}\label{eq:forward-invariance}
    \xB(0) \in \calC \quad\Longrightarrow\quad \xB(t) \in \calC, ~\forall t \in [0, T].
\end{equation}
A widely used tool for certifying forward invariance is the concept of \emph{barrier functions}~\cite{ames2019control}. Specifically, if there exists a continuously differentiable function $h(\xB)$ such that: 
\textit{i)} $h(\xB) \geq 0$ for all $\xB \in \mathcal{C}$ and $h(\xB) < 0$ for all $\xB \notin \mathcal{C}$, and 
\textit{ii)} $\dot{h}(\xB) + \alpha(h(\xB)) \geq 0$ for all $\xB \in \mathbb{R}^d$, where $\alpha(\cdot)$ is a strictly increasing function with $\alpha(0) = 0$,
then the set $\mathcal{C}$ is forward invariant for the dynamical system $\dot{\xB} = \fB(\xB)$. In control theory, forward invariance is closely tied to the notion of \emph{safety}: if a trajectory remains in a predefined safe set, the system is considered safe throughout its evolution.

This notion can be naturally extended to DNNs formulated as continuous-time dynamical systems. By enforcing the forward invariance of specific sets, one can constrain the evolution of the DNN to satisfy desired safety or performance properties. For example, \cite{YangCertifiablyRobustNeural2023} incorporates barrier functions to enforce the forward invariance of robustness sets. By embedding the barrier condition as a penalty term in the loss function, the method enhances the adversarial robustness of Neural ODEs across various datasets. Similarly, \cite{XiaoForwardInvarianceNeural2023} proposes a technique for certifying output specifications using barrier functions, and introduces a novel method for backpropagating invariance constraints through the DNN. These approaches enable Neural ODEs to fulfill behavioral requirements such as avoiding collisions or remaining within safe operational bounds. Extending these ideas, \cite{daiSafeFlowMatching2025} integrates barrier functions into the flow matching framework, introducing a motion planning algorithm that provides safety guarantees.

\textbf{Optimal control.} 
Optimal control is one of the most closely related theoretical frameworks to DNNs when they are interpreted as differential equations. For a dynamical system of the form $\dot{\xB} = \fB(\xB, \uB)$, where $\uB$ denotes the control input, the goal of \emph{optimal control theory}~\cite{athans2013optimal} is to determine a control policy $\uB^*(t)$ that minimizes a prescribed \textit{cost functional} over time. Interestingly, the objective of deep learning can be viewed through the same lens \cite{LiDynamicalSystemsMachine, EMeanfieldOptimalControl2018}. When a DNN is described as a differential equation $\dot{\xB} = \fB_{\thB}(\xB)$, where $\thB$ represents the learnable parameters of the DNN, one can interpret $\thB$ as a form of control signal. From this perspective, training a DNN becomes an instance of solving an optimal control problem, in which the goal is to optimize the parameters (controls) to minimize a loss function (cost).

This connection enables the application of principles from optimal control theory to deep learning. For example, just as the condition $\nabla \fB(\xB^*) = \bm{0}$ characterizes optimality in unconstrained optimization and underpins gradient-based methods, the mean-field Pontryagin’s Maximum Principle (PMP) provides necessary conditions for optimality when training DNNs viewed as dynamical systems. This insight has motivated a variety of control-theoretic training algorithms for DNNs.
For instance, \cite{LiMaximumPrincipleBased2018} proposed a PMP-based training algorithm with theoretical convergence guarantees and enhanced robustness in flat or saddle regions, addressing key limitations of conventional gradient descent. Building on this idea, \cite{LiOptimalControlApproach2018} extended the PMP-based framework to support the training of quantized DNNs. In parallel, \cite{ChenNeuralOrdinaryDifferential2018a} introduced the \emph{adjoint sensitivity method}, which circumvents the need to backpropagate through the internal operations of the ODE solver, thereby improving memory efficiency. In fact, the adjoint sensitivity method is a special case of PMP. More recently, \cite{ZhangYouOnlyPropagate2019} extended this line of research by leveraging PMP to accelerate adversarial training, demonstrating the broader utility of optimal control tools in modern deep learning.

\textbf{Collocation methods.} 
One of the most significant limitations of Neural ODEs is their high computational cost during training~\cite{zhao2025accelerating}. This inefficiency primarily stems from the reliance on ODE solvers, which requires repeatedly evaluating the DNN during the numerical ODE solving process. A promising approach to mitigate this bottleneck is to adopt collocation methods~\cite{alexander1990solving}, a class of numerical techniques for approximating solutions to ODEs. Rather than enforcing the ODE continuously over an interval, collocation methods require the ODE to hold at a discrete set of points, known as \emph{collocation points}.

Leveraging this idea, several recent works have proposed ODE-solver-free training strategies for Neural ODEs to improve computational efficiency. For example, \cite{quaglinoSNODESpectralDiscretization2020} utilized spectral element methods to train Neural ODEs, employing truncated Legendre polynomials~\cite{hussaini1986spectral} to approximate trajectories and regressing the vector field based on estimated derivatives at collocation points. Similarly, \cite{roeschCollocationBasedTraining2021} formulated the training task as an inverse problem~\cite{liang2008parameter}, where the derivatives of the trajectories are first estimated from observed data and then used to fit the underlying vector field. However, both of these methods depend on numerical differentiation at collocation points, which can be sensitive to noise and introduce approximation errors. To address this issue, \cite{zhao2025accelerating} proposed a variational approach inspired by \cite{brunel2014parametric}, which defines a loss function in integral form directly on the collocation points, thus eliminating the need for explicit derivative estimation. This variational formulation improves both robustness to noisy data and training efficiency, making it a compelling alternative to traditional ODE-solver-based training.

\subsection{Advancing DNNs at Layer Level}

We begin by examining how the differential equation perspective can be used to better understand DNNs at the \emph{layer level}. A recent line of work leverages SSMs to better understand the differing behaviors of core deep learning modules. Notably, \cite{sieber2024understanding} propose the Dynamical System Framework (DSF), which recasts attention mechanisms, Deep SSMs, and certain RNNs as time-varying linear SSMs of the form:
\begin{equation}
    \begin{aligned}
        \hB_{i+1} &= \bm{\Lambda}_i \hB_i + \BB \uB_i, \\
        \yB_i &= \CB_i \hB_i + \DB_i \uB_i,
    \end{aligned}
\end{equation}
where $\hB_i$ denotes the hidden state (initialized with $\hB_{-1}=\bm{0}$), $\bm{\Lambda}_i$ is a diagonal state-transition matrix, and $\BB_i$, $\CB_i$, and $\DB_i$ are the input, output, and residual-connection matrices, respectively. Stacking the sequence yields the compact relation
\begin{equation}
    \yB = \bm{\Phi}\uB,
\end{equation}
with $\bm{\Phi}$ determined by the system parameters. This unified representation of different DNNs permits a principled comparison of architectures via classical dynamical-systems tools. Further more, this framework clarifies several empirical phenomena and suggests new design principles. For example, why softmax attention is typically more expressive than linear attention: the former admits a higher-dimensional hidden state, enabling richer dynamics. DSF also motivates architectural modifications, such as substituting the state-transition matrix of a conventional RNN with the Mamba transition used in recent SSMs—which empirically improve RNN performance.

Beyond DSF, several works adopt alternative theoretical tools to analyze the capabilities and limitations of Deep SSMs. For instance, \cite{CironeTheoreticalFoundationsDeep2024} reinterpret S4 and Mamba architectures as linear controlled differential equations (CDEs), enabling the use of \emph{signature expansion} from rough path theory~\cite{friz2020course} to rigorously assess their expressivity, particularly the benefits conferred by the selection mechanisms embedded in modern SSMs. In another direction, \cite{zubicRegularityStabilityProperties2025} investigate the stability and regularity properties of Deep SSMs using concepts from \emph{passivity theory} and \emph{Input-to-State Stability} (ISS). They establish that the inherent energy dissipation in these models leads to an ``exponential forgetting of past states'': in the absence of input, the system’s state decays exponentially, thereby mitigating the influence of initial conditions. Furthermore, they show that the unforced dynamics admit a minimal quadratic energy function whose associated matrix satisfies a property called ``robust AUCloc regularity.'' This structural regularity ensures that the system remains well-behaved even under discontinuous gating mechanisms, such as those introduced by abrupt changes in internal parameters. \cite{NishikawaStateSpaceModels2024} employ $\gamma$-smooth and piecewise $\gamma$-smooth function classes to examine the estimation power of Deep SSMs, concluding that SSMs may match Transformers in estimation capability. \cite{HalloranMambaStateSpaceModels2024} uses Lyapunov exponents to demonstrate that Mamba SSMs are significantly more stable to changes introduced by mixed-precision than comparable Transformers, then further 

We now examine how incorporating domain-specific insights from differential equations can enhance the performance of Deep SSMs. \cite{ParnichkunStateFreeInferenceStateSpace2024} leverages the transfer-function (frequency-domain) representation of SSMs to devise a \emph{state-free} inference scheme whose memory and computational requirements remain essentially constant as the latent-state dimension grows. Because the transfer function can be evaluated with a single Fast Fourier Transform, their method accelerates S4 training substantially without sacrificing state-of-the-art accuracy on downstream tasks. Building on the close connection between CNNs and Deep SSMs, \cite{li2023what} analyzes the convolutional kernels implicitly induced by S4. They show that high-performing kernels scale sub-linearly with sequence length and exhibit a natural decay in weights, assigning higher importance to nearby tokens than to distant ones. Guided by this observation, they introduce \emph{Structured Global Convolution}, a variant that preserves the expressiveness of S4 while delivering faster inference and strong empirical results across multiple tasks. 

\section{Real-world Applications}\label{sec:application}

In this section, we aim to answer the \textbf{third key question}: \emph{What real-world applications benefit from grounding DNNs in differential equations?} We organize the discussion around three broad application areas: time series tasks, generative AI, and large language models.

\subsection{Time Series Tasks}

One of the most significant advantages of the differential equation perspective in deep learning lies in its ability to model time series data. On one hand, differential equations are foundational tools in applied mathematics and are widely used to describe dynamical systems across various scientific domains. In many real-world scenarios, the evolution of time series data is governed by underlying physical laws, which are naturally expressed in differential form. In this context, neural differential equations provide a principled framework for integrating such prior knowledge, offering a physics-informed machine learning approach that often leads to improved performance.

On the other hand, neural differential equations offer a natural solution for handling \emph{irregularly sampled} time series, which are commonly encountered in practice. For instance, medical time series often contain irregular sampling due to patients visiting hospitals only when symptoms arise. Likewise, in cyber-physical systems, sensor malfunctions or adverse environmental conditions can result in missing or asynchronous measurements. Traditional sequence models such as RNNs and Transformers typically assume fixed sampling intervals and lack mechanisms to represent time as a continuous variable, limiting their ability to capture irregular temporal dynamics. In contrast, the differential equation framework, particularly neural differential equations, naturally supports continuous-time modeling. Temporal structure is inherently captured by the differential operator, allowing these models to flexibly adapt to non-uniform sampling rates. These properties make neural differential equations especially well-suited for modeling irregularly sampled time series across a wide range of domains. In the following, we present specific applications that demonstrate the effectiveness of this paradigm. A summarizing table on time series applications is presented in Tab.~\ref{tab:time-series-applications}.

\begin{table}[htbp]
    \centering
    \caption{Applications of differential equation perspective of DNNs on time series tasks.}
    \label{tab:time-series-applications}
    \begin{tabular}{c|c|c}
    \toprule
    \textbf{Fields} &  \textbf{Detailed Fields} & \textbf{Works} \\ \midrule
    \multirow{5}{*}{\raisebox{-0.7\totalheight}{%
      \shortstack{Natural\\Science}
    }}
    & Climate modeling
    & \cite{verma2024climode, ramadhan2020capturing}
    \\
    & Hydrological systems
    & \cite{quaghebeur2022hybrid, huang2025training}
    \\
    & Chemistry
    & \cite{owoyele2022chemnode, yin2023generalized}
    \\
    & Astrophysics
    & \cite{lanzieri2022hybrid}
    \\
    & Turbulence forecasting
    & \cite{portwood2019turbulence}
    \\ \midrule

    \multirow{2}{*}{\raisebox{-0.3\totalheight}{%
      \shortstack{Engineering}
    }}
    & Cyber-Physical Systems
    & \cite{aryal2023application, taboga2024neural}
    \\
    & Robotics
    & \cite{wen2022social, chahine2023robust, duong21hamiltonian}
    \\ \midrule

    \multirow{4}{*}{\raisebox{-0.4\totalheight}{%
      \shortstack{Medical\\Applications}
    }}
    & PhysioNet Challenge
    & \cite{rubanova2019latent, KidgerNeuralControlledDifferential2020}
    \\
    & Clinical pharmacology
    & \cite{lu2021neural, hess2023bayesian, geng2017prediction}
    \\
    & Epidemiology
    & \cite{zhao2025accelerating}
    \\
    & Medical image analysis
    & \cite{niu2024applications}
    \\
    \bottomrule
    \end{tabular}
\end{table}

\textbf{Natural Science.} Given that many natural phenomena are governed by differential equations, neural differential equations have found broad applications across a variety of scientific domains, including climate modeling \cite{verma2024climode, ramadhan2020capturing}, hydrological systems \cite{quaghebeur2022hybrid, huang2025training}, chemistry \cite{owoyele2022chemnode, yin2023generalized}, astrophysics \cite{lanzieri2022hybrid}, and turbulence forecasting \cite{portwood2019turbulence}. In these contexts, neural differential equations are often employed in a hybrid modeling framework, where known physical principles are incorporated into the vector fields of the differential equations, while deep neural networks are used to approximate unknown components such as stochastic effects or unmodeled dynamics.

For example, ClimODE \cite{verma2024climode} proposed a continuous-time neural advection PDE framework for climate and weather modeling by deriving a tailored ODE system for numerical weather prediction. This approach achieved strong performance in forecasting both global and regional weather dynamics. In astrophysics, \cite{lanzieri2022hybrid} embedded domain-specific physical knowledge into Neural ODEs to enable efficient and accurate N-body simulations. Likewise, \cite{portwood2019turbulence} used Neural ODEs to model the transient dissipation of turbulent kinetic energy, effectively capturing key dynamic patterns in turbulent flows.

\begin{casestudy}{ClimODE \cite{verma2024climode}}
    From first principles, weather systems can be modeled as fluxes, representing the spatial transport of physical quantities over time. This process is governed by a PDE of the form:
    \begin{equation}\label{eq:climate-pde}
        \underbrace{\dv{u}{t}}_{\text{time evolution } \dot{u}} + \quad \underbrace{\overbrace{\vB \cdot \grad u}^{\text{transport}} + \overbrace{u\grad \cdot \vB}^{\text{compression}}}_{\text{advection}} = \underbrace{s}_{\text{sources}},
    \end{equation}
    where $u(\xB, t)$ denotes a physical quantity (e.g., temperature) evolving in space $\xB$ and time $t$,  $\vB(\xB, t)$ is the velocity of the flow, and $s(\xB, t)$ represents external sources. Although such formulations are theoretically well-founded, numerically solving this PDE over large-scale domains is computationally expensive. Moreover, deriving accurate closed-form expressions for each term in complex, multiscale systems is often infeasible.

    To address these challenges, ClimODE proposes a continuous-time neural advection PDE framework for climate and weather modeling. Specifically, ClimODE uses a deep neural network to model the temporal evolution of the velocity field: 
    \begin{equation}
        \dot{\vB}(\xB, t) = \fB_{\thB}(u(t), \grad u(t), \vB(t), \psi),
    \end{equation}
    where $\psi$ denotes spatiotemporal embeddings. Substituting this into the original PDE yields a coupled system of first-order ODEs:
    \begin{equation}
        \begin{bmatrix} u(t) \\ \vB(t) \end{bmatrix} = \begin{bmatrix} u(t_0) \\ \vB(t_0) \end{bmatrix} + \int_{t_0}^{t} \begin{bmatrix} -\grad \cdot (u(\tau) \vB(\tau)) \\ \fB_{\thB}(\uB(\tau), \grad \uB(\tau), \vB(\tau), \psi) \end{bmatrix} \dd{\tau}.
    \end{equation}
    This reformulation allows the model to learn complex spatiotemporal dynamics while leveraging the structure of physical conservation laws.
\end{casestudy}

\textbf{Engineering.}
In engineering domains, neural differential equations are frequently employed for system identification tasks. For instance, \cite{rahman2022neural} evaluated Neural ODEs on a suite of benchmark problems commonly used in control engineering, demonstrating strong identification performance. In the context of Cyber-Physical Systems, \cite{aryal2023application} applied Neural ODEs to model and infer critical state variables of power system frequency dynamics. Similarly, \cite{taboga2024neural} used Neural ODEs to predict both temperature and HVAC power consumption. These predictions were integrated into a planning algorithm within a model predictive control framework to adjust temperature setpoints, thereby reducing HVAC energy usage and improving compliance with demand response programs.

Neural ODEs have also seen increasing adoption in robotics, where modeling continuous-time dynamics is crucial. For instance, \cite{wen2022social} proposed Social ODE, which leverages Latent ODEs to forecast multi-agent trajectories in dynamic and socially interactive environments. In the context of drone navigation, \cite{chahine2023robust} introduced Liquid Time-Constant (LTC) networks, a biologically inspired variant of Neural ODEs, to enable robust policy learning. The causal reasoning capabilities inherent in LTC networks allow drones to adapt effectively to significant environmental changes, thereby improving navigation reliability. Additionally, \cite{duong21hamiltonian} presented a Hamiltonian formulation of Neural ODEs on the SE(3) manifold to approximate the dynamics of rigid-body systems. This approach generalizes to various platforms, including pendulums, rigid-body mechanisms, and quadrotors.

\begin{casestudy}{Liquid Neural Networks\cite{chahine2023robust}}
    This work introduces a specialized class of neural networks, referred to as liquid neural networks, designed for vision-based fly-to-target tasks in autonomous drone navigation. The core building block of these networks is the liquid time-constant (LTC) network \cite{HasaniLiquidTimeconstantNetworks2021}, which constitutes a particular form of Neural ODEs:
    \begin{equation}
        \dot{\xB}(t) = -\left[
            \frac{1}{\tau} + \fB(\xB, \IB, t, \thB)
        \right] \odot \xB(t) + \fB(\xB, \IB, t, \thB),
    \end{equation}
    where $\tau$ is a learnable time constant. On one hand, this formulation is loosely inspired by computational models of neural dynamics observed in small biological organisms, incorporating elements of synaptic transmission. On the other hand, it bears structural similarity to Dynamic Causal Models (DCMs) \cite{friston2003dynamic}, which have been successfully used to model complex time-series data such as fMRI signals. These characteristics endow liquid neural networks with strong causal reasoning capabilities, allowing them to generalize effectively across varying conditions.

    Leveraging this causal structure, liquid neural networks have demonstrated state-of-the-art performance in vision-based drone navigation. Notably, they exhibit robust generalization even under significant environmental changes -- for instance, when transitioning from forested regions to urban landscapes -- while relying purely on visual input.
\end{casestudy}

\textbf{Medical Applications.}
As one of the most fundamental tools in applied mathematics, differential equations have long been utilized in medical applications—even prior to the rise of neural differential equations. For example, they have been used to model neuronal activity \cite{breakspear2017dynamic} and tumor growth dynamics \cite{sharma2016analysis}. In this context, neural differential equations have emerged as powerful tools for modeling complex medical time series since their initial introduction. Notably, Latent ODEs \cite{rubanova2019latent} and Neural CDEs \cite{KidgerNeuralControlledDifferential2020} were evaluated on the PhysioNet datasets \cite{reyna2020early, goldberger2000physiobank, silva2012predicting}, which contain irregularly sampled time series representing the clinical trajectories of patients in intensive care units (ICUs). The core task is binary classification of patient mortality, where both models achieved strong empirical performance.

In clinical pharmacology, \cite{lu2021neural} applied Neural ODEs to forecast patient-specific pharmacokinetics, demonstrating substantially improved accuracy in predicting responses to new treatment regimens. To address uncertainty quantification, \cite{hess2023bayesian} proposed Bayesian Neural CDEs for continuous-time treatment effect estimation—marking the first tailored neural framework for providing uncertainty-aware estimates. Their model achieved excellent results on established tumor growth datasets based on pharmacokinetic-pharmacodynamic modeling \cite{geng2017prediction}. 

In epidemiology, Neural ODEs have also proven effective for modeling population dynamics across different groups. For instance, \cite{zhao2025accelerating} applied Neural ODEs to real-world COVID-19 data, demonstrating high accuracy in forecasting infection trends and progression across heterogeneous populations.

In addition, Neural CDEs have also been applied to multimodal settings. For example, \cite{cheruvu2023application} used Neural CDEs to model the progression of pulmonary fibrosis by integrating irregularly sampled structural and imaging data, outperforming LSTM-based baselines. In addition, \cite{niu2024applications} provided a comprehensive survey on the application of Neural ODEs in medical image analysis, covering a wide range of tasks including segmentation, reconstruction, registration, disease prediction, and data generation.


\subsection{Generative AI}

Flow-based generative models, as discussed in Sec.~\ref{subsec:generative}, have achieved state-of-the-art performance and are being widely adopted across a broad range of application domains. In the following, we highlight several key areas where these models have made significant impact, along with representative works in each domain.

\textbf{Computer Vision.}
One of the most prominent application areas is \emph{image generation}, where numerous representative models (e.g., DDPM~\cite{ho2020denoising}, Latent Diffusion Model (LDM)~\cite{rombach2022high}, Flow Matching~\cite{LipmanFlowMatchingGenerative2022}, and Rectified Flow~\cite{liu2023flow}, etc.) were initially developed. Building upon this, \emph{video generation} has emerged as a natural extension, with methods like~\cite{khachatryan2023text2video, gupta2024photorealistic} demonstrating the scalability of flow-based models to high-dimensional, temporally coherent outputs. These models are now widely used in industry, with notable examples including \emph{Stable Diffusion} for text-to-image synthesis and \emph{Sora} for advanced text-to-video generation, underscoring the maturity and real-world relevance of flow-based generative models in vision tasks.

\textbf{Text Generation.}
While autoregressive language models~\cite{touvron2023llama, achiam2023gpt} currently dominate text generation, flow-based and diffusion-based models have emerged as promising alternatives due to their potential for parallel generation. For example, \cite{austin2021structured} proposed a structured denoising diffusion model for discrete sequences and evaluated it on text generation tasks. \cite{gat2024discrete} adapted flow matching for discrete data, demonstrating applications in both text and code generation. \cite{xu2024energy} introduced the Energy-based Diffusion Language Model (EDLM), which uses energy-based modeling over full sequences at each diffusion step to capture token dependencies. \cite{sahoo2024simple} showed that masked diffusion language modeling can be competitive when combined with a simplified Rao-Blackwellized objective and proper training strategies. Most recently, \cite{nie2025large} introduced LLaDA, a large-scale diffusion model that achieves performance comparable to LLaMA3-8B. For a more comprehensive overview, we refer readers to the survey by~\cite{li2025survey}.


\textbf{Protein Design.}  
In the domain of biological sequence and structure generation, flow-based models have enabled scalable protein design. 
\cite{watson2023novo} introduced \emph{RoseTTAFold Diffusion (RFdiffusion)}, which fine-tunes a structure prediction model for denoising-based backbone generation, enabling the design of diverse proteins from simple molecular specifications. 
\cite{jendrusch2025efficient} proposed \emph{Salad}, a family of sparse all-atom denoising models capable of generating structures up to 1,000 amino acids. They also introduced a \emph{structure editing} method to extend the model's applicability to unseen design tasks, such as motif scaffolding and multi-state protein design.


\textbf{Robotics.}
Flow and diffusion models have also been increasingly adopted in robotic policy learning. 
\cite{chi2025diffusion} introduced \emph{Diffusion Policy}, which formulates visuomotor policy learning as a conditional denoising diffusion process. 
In addition to training policies from scratch, flow-based models are widely used in vision-language-action (VLA) frameworks to model complex action distributions. Examples include RDT-1B~\cite{liu2024rdt}, HybridVLA~\cite{liu2025hybridvla}, DexVLA~\cite{wen2025dexvla}, $\pi_0$~\cite{black2024pi_0}, $\pi_{0.5}$~\cite{intelligence2025pi_}, and $\pi_{0.6}^*$ \cite{amin2025pi}. A broader overview can be found in the survey by~\cite{wolf2025diffusion}.

\begin{casestudy}{Vision-Language-Action models \cite{black2024pi_0, intelligence2025pi_, amin2025pi}}
    In recent years, Vision–Language–Action (VLA) models have emerged as foundational techniques in embodied AI. These models leverage pretrained Vision–Language Models (VLMs) to transfer high-level reasoning and world knowledge to robotic control. Among state-of-the-art VLAs, the $\pi$ series (including $\pi_0$~\cite{black2024pi_0}, $\pi_{0.5}$~\cite{intelligence2025pi_}, and $\pi_{0.6}^*$~\cite{amin2025pi}) has established \emph{flow matching} as a core architectural component. Early VLAs, such as OpenVLA~\cite{kim2024openvla}, directly repurpose the least-used 256 tokens as action tokens and generate robot actions in an autoregressive manner. Although straightforward, autoregressive inference suffers from limited throughput, making it inadequate for high-frequency or highly dexterous control tasks that require rapid responses and precise motion execution. In contrast, the $\pi$-series VLAs employ flow matching as an action expert to model complex action distributions. This formulation enables the generation of \emph{multi-step action chunks} within a single inference pass, substantially improving control frequency and strengthening planning capabilities relative to autoregressive approaches.

\end{casestudy}

\subsection{Large Language Model Alignment}\label{subsec:llm-align}

In recent years, autoregressive large language models (LLMs) such as ChatGPT \cite{achiam2023gpt}, LLaMA \cite{touvron2023llama}, and DeepSeek \cite{liu2024deepseek} have emerged as powerful tools, demonstrating exceptional performance across a wide array of tasks. However, despite their impressive capabilities, these models continue to face substantial challenges. Their reliance on massive training datasets, which are often scraped from the web with minimal curation, introduces inherent errors, biases, and inconsistencies. Consequently, LLMs may confidently produce fabricated "facts," reinforce embedded societal prejudices, or generate content that is offensive, insensitive, or contextually inappropriate.

To address these concerns, alignment strategies have become crucial for guiding LLMs toward desired behaviors. The main alignment techniques include (i) prompt engineering, (ii) fine-tuning, and (iii) representation engineering. A particularly promising advancement is the differential equation perspective, providing a unified theoretical framework that interprets and enhances these methods through principles from control theory. By modeling LLMs as stochastic dynamical systems, alignment tasks can naturally be formulated as control problems.

\textbf{Prompt Engineering.} 
Prompt engineering aims to steer the outputs of LLMs using carefully designed prompts. In this context, we can view these prompts as \textit{control inputs}. Under this perspective, \cite{bhargava2023s, soatto2023taming} examine the controllability of LLMs. Similarly, \cite{luo2023prompt} demonstrates the equivalence between multi-round prompt engineering and optimal control tasks, offering a comprehensive survey of existing literature within this unified framework.

\textbf{Representation Engineering.}
In representation engineering, perturbations introduced to steer model outputs are treated explicitly as control inputs, often computed through optimal control formulations. For example, RE-Control \cite{kong2024aligning} formulated the alignment objective as the maximization of a value function, represented by a three-layer MLP trained to predict expected rewards, with optimal perturbations calculated via gradient ascent. Conversely, LiSeCo \cite{cheng2024linearly} emphasized minimizing perturbation magnitudes subject to semantic constraints, highlighting different practical approaches within the representation engineering paradigm. More recently, ODESteer \cite{zhao2026odesteer} proposed a unified ODE-based framework for representation engineering methods. Specifically, it interprets the widely used activation addition technique as an Euler discretization of an ODE, and on this basis uses barrier functions from control theory to identify steering directions.

\begin{casestudy}{ODESteer \cite{zhao2026odesteer}}
    ODESteer provides a unified framework for representation engineering methods for LLM alignment. Its central insight is that the widely used technique of \textit{activation addition} admits a natural interpretation in terms of ODEs. Specifically, activation addition can be written as
    \begin{equation}\label{eq:act-add}
        \tb{a} = \aB + T \cdot \vB(\aB),
    \end{equation}
    where $\tb{a}$ denotes the steered activation, $\vB(\aB)$ is the steering vector, and $T$ is a scalar that controls the intervention strength. This formulation is precisely the Euler discretization of the ODE $\db{a}(t) = \vB(\aB(t))$. In particular, applying one Euler step from $t=0$ to $t=T$ gives
    \begin{equation}\label{eq:act-ode}
        \aB(T) = \aB(0) + \db{a}(0)\cdot (T - 0) = \aB(0) + T \cdot \vB(\aB(0)).
    \end{equation}
    This expression coincides with Eq.~\eqref{eq:act-add} when $\aB(T)$ is identified with the steered activation $\tb{a}$. Therefore, standard activation addition can be viewed as taking a single Euler step from $\aB(0)$ with step size $T$.
    
    Building on this connection, the authors showed that two major classes of representation engineering methods, namely \textit{input reading} and \textit{output optimization}, can be unified through the notion of a barrier function that determines the steering direction. From this perspective, they further proposed a new representation engineering method derived from the unified framework. In particular, by defining the barrier function as the \textit{log-density ratio} between the positive and negative activation distributions, they developed a multi-step, adaptive representation engineering approach.
\end{casestudy}

\begin{remark}
    In most existing works that integrate LLMs with control theory, LLMs are typically treated as discrete dynamical systems, rather than continuous-time systems governed by differential equations. However, we argue that this integration can -- and should -- be interpreted through the lens of continuous dynamical systems. This is because discrete systems often arise as numerical approximations of continuous ones, and many control-theoretic methods applied in these contexts are originally developed in the continuous domain and later adapted to discrete settings. We hope that this differential equation perspective will motivate future research to explore more direct and principled connections between continuous-time control theory and LLMs.
\end{remark}

\section{Challenges and Opportunities}\label{sec:discussion}

Although the differential equation perspective offers a principled foundation for understanding, analyzing, and improving DNNs, turning this perspective into a more systematic theory of DNNs still requires addressing important practical and conceptual gaps. In this section, we highlight several major challenges and promising future directions.

\textbf{Computational Efficiency and Scalability.}
One of the principal challenges in interpreting DNNs through the lens of continuous-time dynamical systems lies in computational efficiency and scalability. Although this perspective provides valuable theoretical insights, its practical application necessitates discretization, and the computational burden is highly sensitive to the choice of discretization scheme. In particular, employing standard ODE solvers during training can be prohibitively expensive, as the solver must repeatedly evaluate the DNN’s vector field. This problem is exacerbated in the presence of \emph{stiff} dynamics, where rapid state variations require very small integration steps to ensure numerical stability. Consequently, reliance on continuous-time solvers introduces variability in computational cost and poses serious limitations for large-scale, high-dimensional or real-time applications.

\textbf{Disconnect Between Model-Level and Layer-Level Perspectives.}
Another major difficulty is the lack of a unified framework bridging the model-level and layer-level differential equation viewpoints. Model-level approaches typically characterize the global evolution of feature representations by treating the entire DNN as a single dynamical system. In contrast, layer-level formulations focus on dynamics of a specific component of DNNs, which often approximates each layer as a discretized operator derived from a simple ODE. Despite progress in both directions, the absence of systematic methodologies for transferring insights across these levels hinders generalization and limits practical applicability. Bridging this divide may require the incorporation of more advanced mathematical tools, such as PDEs, rough path theory, or operator-theoretic techniques.

\textbf{Interpretability.} 
Despite the theoretical appeal of differential equation-based analyses, DNNs remain largely opaque in practice. Continuous-time formulations may expose global dynamical trends, but they often fail to illuminate the inner workings of specific components, such as individual neurons or submodules. This interpretability gap limits the utility of differential equation tools in tasks that demand transparency or formal guarantees. Developing frameworks that not only model DNNs as dynamical systems but also yield actionable insights into their structure and behavior remains an open and pressing challenge.

\textbf{Integration with Emerging AI Paradigms.} 
While differential equation-based techniques have been successfully applied to various deep learning problems, their integration with emerging AI paradigms, particularly LLMs, remains relatively underexplored. As discussed in Section \ref{subsec:llm-align}, recent studies have demonstrated that concepts from control theory and dynamical systems can inform areas such as prompt engineering and representation learning, especially within alignment frameworks. These insights could potentially be extended to other challenging domains, such as multi-step reasoning or structured decision-making. Unlocking these opportunities requires a deeper theoretical and algorithmic synthesis between continuous-time models and the unique architectural and training features of modern AI systems.

\section{Conclusion}\label{sec:conclusion}

In this work, we systematically examined the mathematical foundations of DNNs through the lens of differential equations. Our analysis considered two complementary levels of abstraction: the \textit{model level}, which interprets the entire network as a continuous-time dynamical system, and the \textit{layer level}, which views individual layers as SSMs. From these perspectives, we demonstrated how differential equations can enhance our understanding of DNN design principles, shed light on theoretical properties such as stability and expressivity, and guide the development of more robust and efficient architectures. In addition, we discussed emerging opportunities for applying differential equation-based frameworks across diverse DNN paradigms. This unified viewpoint underscores the analytical power and practical relevance of integrating differential equations into the study and advancement of modern deep learning.

\bibliographystyle{abbrv}
\bibliography{ref.bib}

\newpage
\appendix
\section{Abbreviation}\label{app:abbrv}

The abbreviations used throughout this survey are compiled and presented in alphabetical order in the table below.

\begin{table}[htbp]
    \centering
    \caption{Summary of abbreviation used in the survey.}
    \label{tab:abbrv}
    \begin{tabular}{c|c}
    \toprule
    \textbf{Abbreviation} & \textbf{Meaning} \\ \midrule
    CDE & Controlled Differential Equation \\
    CNN & Convolutional Neural Network \\
    DNN & Deep Neural Networks \\
    GAN & Generative Adversarial Network \\
    LLM & Large Language Model \\
    LSSL & Linear State-Space Layer \\
    LTI & Linear Time Invariant \\
    NDE & Neural Differential Equation \\
    ODE & Ordinary Differential Equation \\
    PDE & Partial Differential Equation \\
    RDE & Rough Differential Equation \\
    RNN & Recurrent Neural Network \\
    SDE & Stochastic Differential Equation \\
    SSD & State Space Duality \\
    SSM & State Space Model \\
    ZOH & Zero-Order Hold \\
    \bottomrule
    \end{tabular}
\end{table}

\end{document}